\documentclass[lettersize,journal]{IEEEtran}
\usepackage{amsmath,amsfonts,amssymb}
\usepackage{array}
\usepackage[caption=false,font=scriptsize,labelfont=sf,textfont=sf]{subfig}
\usepackage{textcomp}
\usepackage{stfloats}
\usepackage{url}
\usepackage{verbatim}
\usepackage{graphicx}
\usepackage{color}
\usepackage{tikz}
\usepackage{comment}

\usepackage[ruled,vlined]{algorithm2e}
\usepackage{multirow}
 
\usepackage{booktabs}
\usepackage{bm}
\usepackage{dsfont}

\usepackage{hyperref}
\hypersetup{
	colorlinks=true,
	citecolor=cyan,
}

\usepackage{threeparttable}

\begin{document}

\title{\huge RepQuant: Towards Accurate Post-Training Quantization of Large Transformer Models via Scale Reparameterization}

\author{Zhikai Li, \IEEEmembership{Graduate Student Member, IEEE}, Xuewen Liu, Jing Zhang, and Qingyi Gu, \IEEEmembership{Senior Member, IEEE}  
	\thanks{This work was supported in part by the National Natural Science Foundation of China under Grant 62276255; in part by the National Key Research and Development Program of China under Grant 2022ZD0119402. \emph{(Corresponding author: Qingyi Gu.)}}
	\thanks{
		Z. Li, X. Liu, J. Zhang, and Q. Gu are with the Institute of Automation, Chinese Academy of Sciences, Beijing 100190, China, and Z. Li, X. Liu, and J. Zhang are also with the School of Artificial Intelligence, University of Chinese Academy of Sciences, Beijing 100049, China (e-mail: lizhikai2020@ia.ac.cn; liuxuewen2023@ia.ac.cn; zhangjing2024@ia.ac.cn; qingyi.gu@ia.ac.cn).
	}
}



\maketitle

\begin{abstract}
Large transformer models have demonstrated remarkable success across vision, language, and multi-modal domains. However, due to their complex structures, they are criticised for high latency and resource consumption during inference. Post-training quantization (PTQ), which requires only a small dataset for calibration and avoids end-to-end retraining, is a promising solution for compressing these large models. Regrettably, existing PTQ methods typically exhibit non-trivial performance loss, especially in low-bit settings. 
We find that the performance bottleneck stems from over-consideration of hardware compatibility in the quantization process, compelling them to reluctantly employ simple quantizers, albeit at the expense of accuracy.
With the above insights, we propose RepQuant, a novel PTQ framework with quantization-inference decoupling paradigm to address the above issues, and it is compatible with multiple transformer variants. RepQuant employs complex quantizers in the quantization process and simplified quantizers in the inference process, and performs mathematically equivalent transformations between the two through quantization scale reparameterization, thus ensuring both accurate quantization and efficient inference.
More specifically, we focus on two components with extreme distributions: LayerNorm activations with severe inter-channel variations and Softmax activations with power-law characteristics. Initially, we apply channel-wise quantization and log$\sqrt{2}$ quantization, respectively, which are tailored to their distributions. In particular, for the former, we introduce a learnable per-channel dual clipping scheme, which is designed to efficiently identify outliers in the unbalanced activations with fine granularity. Then, we reparameterize the scales to hardware-friendly layer-wise quantization and log2 quantization for inference.
Moreover, quantized weight reconstruction is seamlessly integrated into the above procedure to further push the performance limits. 
Extensive experiments are performed on different large-scale transformer variants on multiple tasks, including vision, language, and multi-modal transformers, and RepQuant encouragingly demonstrates significant performance advantages and strong robustness.
\end{abstract}


\section{Introduction}
\IEEEPARstart{T}HANKS to large-scale modeling, transformer-based models have showcased remarkable prowess across a variety of tasks, such as visual recognition~\cite{han2022survey,dosovitskiy2020image,carion2020end}, language generation~\cite{zhao2023survey,zhang2022opt,touvron2023llama}, and image-text matching~\cite{xu2023multimodal,radford2021learning,kim2021vilt}, etc. 
Nevertheless, this improved performance is accompanied by a substantial increase in the number of model parameters, even to the order of hundreds of billions, which leads to enormous time costs and resource overheads during inference~\cite{wang2022quantformer,tao2022compression,li2023qft,frantar2023sparsegpt}, limiting their potential for widespread real-world applications. As a result, techniques for compressing these large models to improve inference efficiency have received increasing attention.

\begin{figure}
	\centering
	\includegraphics[width=0.9\linewidth]{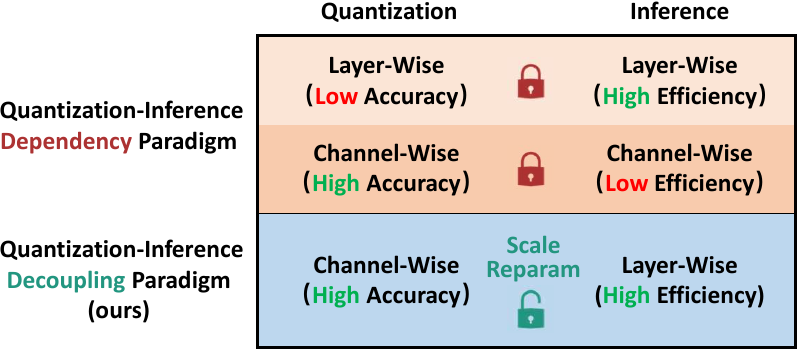}
	\caption{Comparison of different paradigms, with an example of quantization granularity for LayerNorm activations. Our proposed quantization-inference decoupling paradigm shows significant advantages.}
	\label{fig:paradigm}
\end{figure}

Model quantization, which reduces the numerical bit precision of the parameters, is regarded as one of the most effective model compression approaches~\cite{gholami2021survey,krishnamoorthi2018quantizing,wang2022optimization}.
To mitigate the performance degradation, early efforts focus on the end-to-end retraining scheme for the quantized models~\cite{zhou2016dorefa,jacob2018quantization}, referred to as quantization-aware training (QAT). Despite its good performance, such retraining is computationally intensive, which can incur huge resource consumption and delay the production cycle, especially when dealing with large transformer models~\cite{liu2023llm,li2023vit}. In contrast, post-training quantization (PTQ), which utilizes a small unlabeled dataset to calibrate quantization parameters, can overcome the above challenges~\cite{nagel2019data,nagel2020up}. For example, quantizing a model with 175 billion parameters takes only approximately four hours on a single NVIDIA A100 GPU~\cite{frantar2022gptq}. Thus, for large-scale models, PTQ is a more practical and promising solution.

However, unlike convolutional neural networks (CNNs), the activations of transformer models exhibit prominent extreme distributions with systematic outliers, especially in several unique components such as LayerNorm and Softmax, as illustrated in Figs. \ref{fig:layernorm} and \ref{fig:softmax}, which makes quantizing transformer models challenging~\cite{xiao2023smoothquant,yuan2023rptq}.
To this end, various PTQ schemes have been developed to deal with the extreme distributions. For instance, for vision transformers, FQ-ViT~\cite{lin2021fq} proposes powers-of-two scale quantization for LayerNorm activations and PTQ4ViT~\cite{yuan2021ptq4vit} proposes twin uniform quantization for Softmax activations; for language transformers, SmoothQuant~\cite{xiao2023smoothquant} smooths the distribution of LayerNorm activations before quantization. Unfortunately, when performing low-bit (e.g., 4-bit) quantization, these schemes tend to underperform significantly and may even crash.
We thoroughly investigate the performance bottleneck and find that they all consistently adhere to a seemingly intuitive but flawed view that in order to ensure the inference efficiency of the quantized model, they have to carefully employ simple quantizers that match the characteristics of the target hardware in the quantization process, despite the fact that the use of simple quantizers inevitably introduces large quantization bias.

\begin{figure*}
	\centering
	\includegraphics[width=0.98\linewidth]{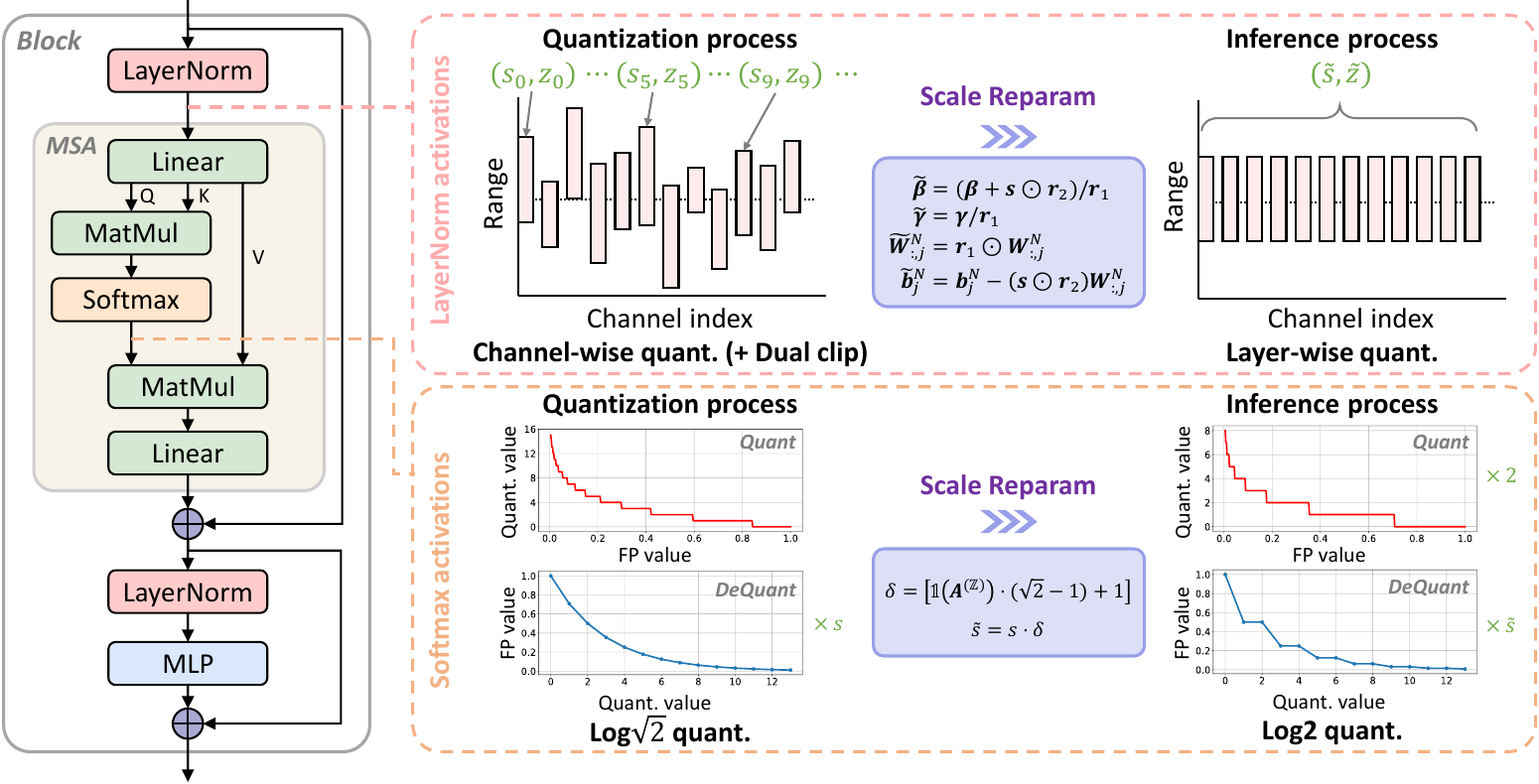}
	\caption{Overview of the proposed RepQuant framework. Based on the quantization-inference decoupling paradigm, we initially apply channel-wise quantization for LayerNorm activations with severe inter-channel variations and log$\sqrt{2}$ quantization for Softmax activations with power-law characteristics in the quantization process, and then we simplify them to layer-wise quantization and log2 quantization via scale reparameterization, respectively, in the inference process, which can ensure both accurate quantization and efficient inference.}
	\label{fig:overview}
\end{figure*}

In this paper, we revisit the conventional view by questioning \emph{whether the traditional quantization-inference dependency paradigm is the only option}. Motivated by this, we explore the feasibility of decoupling the quantization and inference processes, and reveal an intriguing insight: complex quantizers and hardware standards are not always mutually exclusive; instead, the two can be explicitly bridged via mathematically equivalent transformations, which we call \emph{scale reparameterization}. This revelation potentially ushers in a novel quantization-inference decoupling paradigm, i.e., in the initial quantization process, complex quantizers that are tailored to the extreme distributions are employed to sufficiently reduce quantization bias, and then they are simplified to hardware-friendly quantizers via scale reparameterization for efficient inference, which can ensure both high quantization accuracy and inference efficiency. An explicit comparison of the two paradigms is presented in Fig. \ref{fig:paradigm}.

Consequently, we propose RepQuant, a new PTQ framework for large transformer models that escapes from the traditional paradigm. 
In RepQuant, we first strategically customize the quantizers for the extreme distributions of activations: for LayerNorm activations, we apply channel-wise quantization to accommodate the severe inter-channel variations; and for Softmax activations, we apply log$\sqrt{2}$ quantization to match the power-law characteristics. 
In particular, for channel-wise quantization, we introduce a per-channel dual clipping method to learn the numerical upper and lower bounds of each channel at a fine-grained level, which allows for the efficient and precise identification and elimination of outliers. 
Then, we simplify the above quantizers to layer-wise quantization and log2 quantization by scale reparameterization, respectively, to match the hardware requirements. 
Note that scale reparameterization in RepQuant is an equivalent transformation and is theoretically supported, and thus has the potential to enjoy both the accuracy of complex quantizers and the efficiency of simplified quantizers.
Moreover, we seamlessly integrate quantized weight reconstruction into the above procedures, resulting in a sequential quantization pipeline.
The RepQuant framework has no bells and whistles and thus is compatible with various transformer variants, and its overview is illustrated in Fig. \ref{fig:overview}.

To sum up, our main contributions are as follows:
\begin{itemize}
    \item We revisit the limitations of traditional quantization-inference dependency paradigm and reveal that over-consideration of hardware compatibility during quantization is not always mandatory. This suggests opportunities for accurate quantization of extreme distributions.
    \item We propose a novel PTQ framework for large transformer models, which decouples the quantization and inference processes. It employs complex quantizers for the former and simplified ones for the latter, achieving both high accuracy and efficient inference. The decoupling is bridged by scale reparameterization, an equivalent transformation.
    \item We introduce per-channel dual clipping for channel-wise quantization of activations, which enables fine-grained identification of unbalanced outliers. It minimizes the bias in the actual quantization space, which inherently surpasses methods that smooth outliers in the full-precision space before quantization.
    \item Extensive experiments are performed on various transformer variants on a variety of tasks, including visual recognition, language generation, and image-text matching, and RepQuant significantly outperforms the strong baselines, especially in low-bit cases.
\end{itemize}

Note that this paper extends the preliminary conference paper~\cite{li2023repq}. In this paper, RepQuant is a generic PTQ framework for large transformer models, including vision, language, and multi-modal transformers, while RepQ-ViT in the conference paper can be regarded as a special case for vision transformers. Thus, this paper emphasizes the common problems that challenge existing methods in quantizing various transformers, and that RepQuant consistently provides effective and robust solutions. 
RepQuant further introduces a learnable dual clipping method to achieve accurate and efficient outlier removal in the initial channel-wise quantization of LayerNorm activations, allowing us to minimize the bias in the actual quantization space. This is fundamentally superior to methods that smooth outliers in the full-precision space before quantization, e.g. SmoothQuant, and a detailed comparison and analysis is presented in this paper. Besides, RepQuant seamlessly integrates weight reconstruction and builds a sequential quantization pipeline.
With the above improved modules, RepQuant exhibits strong potential for quantizing transformer models, significantly outperforming RepQ-ViT and existing methods.

\section{Related Works}
\subsection{Large Transformer Models}
Transformer-based models, originally developed for natural language processing tasks~\cite{kenton2019bert}, have undergone significant evolution and blossomed across a spectrum of deep learning applications, including vision, language, and multi-modal domains.
For visual recognition tasks, ViT~\cite{dosovitskiy2020image} successfully applies pure transformer models to the image classification task, achieving competitive results with CNNs.
Swin~\cite{liu2021swin} presents the shifted window attention, boosting the performance  at various scales.
Utilizing ViT~\cite{dosovitskiy2020image} as its image encoder, SAM~\cite{kirillov2023segment} builds a promptable segmentation system, exhibiting exceptional zero-shot generalization capabilities.
Besides the vision domain, language transformers that are scaled to larger sizes, also known as large language models, have revolutionized language generation tasks. Many notable open-source pre-trained models have emerged, such as OPT~\cite{zhang2022opt} and LLaMA~\cite{touvron2023llama,touvron2023llama2}, and they can match the performance of closed competitors like GPT-3~\cite{brown2020language}.
In addition, transformer models also show strong potential in the multi-modal domain. For instance, CLIP~\cite{radford2021learning} combines two transformers into a twin-tower structure, with ViT~\cite{dosovitskiy2020image} as the image encoder and GPT-2~\cite{radford2019language} as the text encoder, to learn the semantic information between images and texts.

However, these models are known to be ``large", with extremely large number of parameters and computations, which makes them suffer from huge memory footprints and slow execution speeds in real-world deployment and inference~\cite{zhu2023survey}. 
For instance, running an OPT model with 175 billion parameters in FP16 requires at least 5 NVIDIA A100 GPUs, with an average per-token latency of 230ms~\cite{frantar2022gptq}. As a result, compressing these models to reduce complexity is highly desired in real-world applications.

\subsection{Post-Training Quantization}
Model quantization, which replaces the original floating-point parameters with low-precision values, is a popular approach to reduce the complexity of neural networks~\cite{gholami2021survey,krishnamoorthi2018quantizing}. As a noteworthy branch, PTQ requires only a small calibration dataset and is free from retraining, and is thus believed to be a promising solution to achieve rapid model compression and deployment~\cite{wang2022optimization}.
There have been several PTQ methods that achieves impressive results on CNNs, such as DFQ~\cite{nagel2019data}, AdaRound~\cite{nagel2020up}, and BRECQ~\cite{li2021brecq}; however, due to the structural differences of transformer models, they typically produce poor performance.
Thus, various efforts have focused on PTQ schemes for transformer models, and we categorize them below by vision and language domains.

\textbf{PTQ for vision transformers}
To mitigate the distributional gap between full-precision and quantized models, ranking loss \cite{liu2021post} is implemented to maintain the consistent ordering of attention scores after quantization.
FQ-ViT \cite{lin2021fq} quantizes LayerNorm and Softmax operations by powers-of-two scale and log-int-Softmax, respectively, to enable full quantization.
PSAQ-ViT \cite{li2022patch,li2023psaq} generates images from pre-trained models, boosting PTQ performance from the perspective of calibration samples.
PTQ4ViT \cite{yuan2021ptq4vit} introduces twin uniform quantization to address the imbalance in the distributions of Softmax and GELU activations, supplemented by a Hessian-based metric to optimize quantization scales. 
APQ-ViT \cite{ding2022towards} attempts to maintain the power-law distributions of Softmax activations and introduces a block-wise calibration strategy that can perceive the overall quantization disturbance.
NoisyQuant~\cite{liu2023noisyquant} reduces the quantization error by actively modifying the activation distribution through additive noise bias, which unfortunately introduces additional computational overhead.

\textbf{PTQ for language transformers}
For large language models, a subset of existing works focuses on weight-only quantization, keeping activations in floating-point values, such as GPTQ~\cite{frantar2022gptq}, AWQ~\cite{lin2023awq}, and SqueezeLLM~\cite{kim2023squeezellm}. 
They are effective in solving the memory-bound problem, but ignore the compute-bound problem, which also strongly affects inference latency~\cite{li2023fptq}. 
The remaining methods work on quantizing both weights and activations, which can address the above issues, but leaves them open to the challenges of extreme distributions of activations~\cite{yao2022zeroquant,wei2022outlier}.
LLM.int8()~\cite{dettmers2022llm} proposes a mixed-precision decomposition scheme to isolate the outlier feature channels separately. However, it is not friendly to hardware implementations and results in slow inference speed.
RPTQ~\cite{yuan2023rptq} reorders the channels and quantizes them in groups to mitigate the effects of inter-channel variations, which also brings additional computation during inference.
SmoothQuant~\cite{xiao2023smoothquant} smooths outliers in activations by manually scaling the range of each channel before quantization.
Based on this, Outlier Suppression+~\cite{wei2023outlier} adds channel-wise shifting to push the performance limits.
Further, OmniQuant~\cite{shao2023omniquant} learns the parameters for the above scaling and shifting, even though the learning process is resource-consuming.

Despite the great progress, the results of the above methods are still far from satisfactory. 
On one hand, most works over-consider hardware compatibility during quantization, leading to inaccurate quantization; on the other hand, a few works introduce additional computational overheads~\cite{liu2023noisyquant,yuan2023rptq} or even require customization of complex operators~\cite{dettmers2022llm}, which ignores hardware efficiency. On the contrary, thanks to the quantization-inference decoupling paradigm, RepQuant allows employing complex quantizers during quantization and simplified quantizers during inference, which can guarantee both accurate quantization and efficient inference.
Notably, the complex quantizers in RepQuant are tailored for extreme distributions and can effectively eliminate outliers, thus potentially enabling the minimization of quantization bias in the actual quantization space, which separates it from the methods that smooth distributions in the full-precision space before quantization~\cite{xiao2023smoothquant,wei2023outlier,shao2023omniquant}.
Moreover, RepQuant exhibits promising generalization and is compatible with new backbone models, such as SAM~\cite{kirillov2023segment} and CLIP~\cite{radford2021learning}, whereas few existing works attempt to quanitize them.


\section{Methodology}
\subsection{Preliminaries}
\subsubsection{Transformer blocks}
The main body of a transformer model is a sequence of stacked blocks.
The encoder blocks in vision transformers and the decoder blocks in language transformers are structurally identical, and they can both be formalized as follows:
\begin{align}
  \bar{\bm{X}} & = \text{MSA}(\text{LayerNorm}(\bm{X})) + \bm{X}, \\
  \bm{Y} & = \text{MLP}(\text{LayerNorm}(\bar{\bm{X}})) + \bar{\bm{X}},
\end{align}
where MSA and MLP are the multi-head self-attention module and multi-layer perceptron module, respectively. MLP is a feed-forward network, and MSA calculates the attention as follows:
\begin{align}
\text{Attn}(\bm{Q},\bm{K},\bm{V}) &= \text{Softmax}\left(\frac{\bm{Q}\cdot \bm{K}^T}{\sqrt{D}}\right)\cdot\bm{V},
\end{align}
where $\bm{Q}$, $\bm{K}$, and $\bm{V}$ are query, key, and value, respectively, and they are obtained by linear projections, i.e., $\bm{Q}=\bm{X'}\bm{W}^q$, $\bm{K}=\bm{X'}\bm{W}^k$, $\bm{V}=\bm{X'}\bm{W}^v$, and $D$ is size of hidden features. Here, $\bm{X'}$ is the LayerNorm activation.

In this paper, we work on quantizing both the inputs and weights of the above matrix multiplications to enable low-precision arithmetic. And we employ the hardware-friendly quantizers below to maximize inference efficiency.

\subsubsection{Efficient quantizers}
The uniform quantizer is the most basic and hardware-friendly quantizer, and it is defined as:
\begin{align}
  Quant&: \bm{x}^{(\mathbb{Z})} = \text{clip}\left(\left\lfloor \frac{\bm{x}}{s} \right\rceil+z, 0, 2^b-1 \right), \\
  DeQuant&: \hat{\bm{x}} = s\left(\bm{x}^{(\mathbb{Z})}-z\right) \approx \bm{x},
\end{align}
where $\bm{x}$ and $\bm{x}^{(\mathbb{Z})}$ are the floating-point and quantized values, respectively, and the de-quantized value $\hat{\bm{x}}$ approximately recovers $\bm{x}$. Here, $\left\lfloor\cdot\right\rceil$ is the round operation, and $b$ is the quantization bit-precision. In PTQ, quantization scale $s\in \mathbb{R}^+$ and zero-point $z \in \mathbb{Z}$ are the most important quantization parameters, which are initially determined by the upper bound $x_{up}$ and lower bound $x_{low}$ as follows:
\begin{equation}
\label{eq:sz}
s = \frac{x_{up}-x_{low}}{2^b-1}, \quad z = \left\lfloor-\frac{x_{low}}{s} \right\rceil.
\end{equation}

The log2 quantizer is another popular choice, which can perform bit-shifting through hardware logic for fast arithmetic and is defined as follows:
\begin{align}
\label{eq:log_quant}
 Quant&: \bm{x}^{(\mathbb{Z})} = \text{clip}\left(\left\lfloor -\log_2 \frac{\bm{x}}{s} \right\rceil, 0, 2^b-1 \right), \\
\label{eq:log_dequant}
  DeQuant&: \hat{\bm{x}} = s\cdot 2^{-\bm{x}^{(\mathbb{Z})}} \approx \bm{x}.
\end{align}

For quantization granularity, applying channel-wise quantization for weights and layer-wise quantization for activations is well supported by compilers and hardware, and thus is already the consensual configuration. Although token-wise dynamic quantization for activations is introduced, it suffers from potential hardware compatibility flaws~\cite{yuan2023rptq,li2023fptq}. Hence, we adopt the consensual configuration in this paper.

\subsection{Challenges of Quantizing Activations in Transformers}
\begin{figure}
	\centering
	\includegraphics[width=0.9\linewidth]{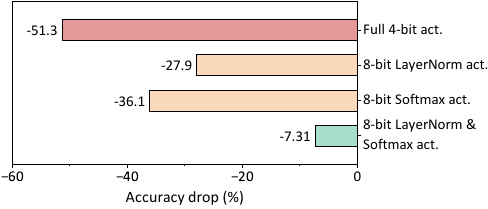}
	\caption{Analysis of performance bottlenecks of quantizing activations in DeiT-S. Evidently, the activations of LayerNorm and Softmax are the most significant obstacles that limit the quantization performance, posing great challenges for low-bit quantization.}
	\label{fig:challenges}
\end{figure}

In contrast to CNNs, due to the introduction of unique components, quantizing activations in transformers is more challenging, and in particular, low-bit quantization suffers from catastrophic performance degradation. To this end, we conduct an experimental analysis to identify the bottlenecks that limit the quantization performance, as shown in Fig. \ref{fig:challenges}.
Here, we take DeiT-S as an example, and all weights are quantized to 4-bit. When performing full 4-bit quantization for activations, the result verges on a crash; whereas reserving the activations of LayerNorm and Softmax as 8-bit, with the rest unchanged, there is a significant performance restoration, narrowing the gap to the full-precision result dramatically. Thus, LayerNorm activations and Softmax activations are the main factors contributing to low quantization performance, and we describe and analyze their distributions in detail below, respectively. 

\subsubsection{Inter-channel variations in LayerNorm activations}
\begin{figure}
	\centering
        \subfloat[DeiT-S]{
	  \includegraphics[width=0.92\linewidth]{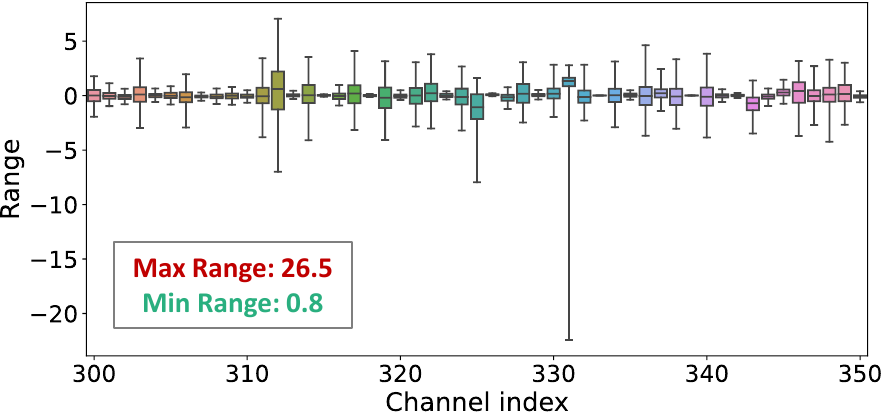}
	} \\
        \subfloat[LLaMA-7B]{
	  \includegraphics[width=0.92\linewidth]{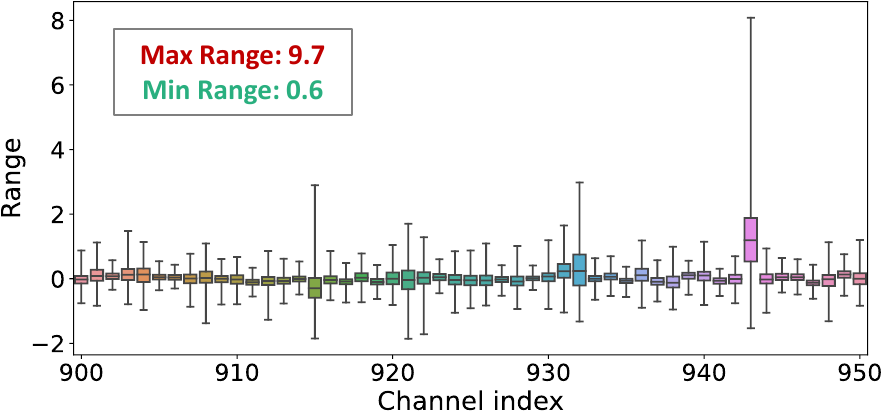}
	}
        
	\caption{Boxplots of different channels of the first module’s LayerNorm activations in DeiT-S and LLaMA-7B. As we can see, the range varies significantly across channels, with tens of times the variation between the maximum and minimum ranges.}
	\label{fig:layernorm}
\end{figure}

We investigate the distributional variations across different channels in LayerNorm activations. 
By visualizing and comparing the ranges of selected channels in DeiT-S and LLaMA-7B models, as shown in Fig. \ref{fig:layernorm}, we clearly demonstrate the existence of severe inter-channel variations.
For instance, in the DeiT-S model, the observed maximum and minimum ranges are markedly different, at 26.5 and 0.8, respectively, which indicates a variation of 33 times between the extreme values. This discrepancy poses a considerable challenge to apply layer-wise quantization, where all channels are forced to adhere to a unified quantization scale, thus inevitably leading to inaccurate and unreliable representations.
Similarly, this issue is also highlighted in the LLaMA-7B model, and due to the larger number of channels (4096), it can potentially be further exacerbated. The use of a single quantization scale in such a context leads to even greater error accumulation and underscores a more pronounced challenge. 

A direct approach to resolving the above problem is to apply channel-wise quantization, i.e., each channel owns its individual quantization scale. However, it necessitates the support of specialized hardware and increases computational burden and inference latency. Consequently, how to accurately represent the distribution with inter-channel variations while ensuring efficient inference remains an open issue.

\subsubsection{Power-law characteristics in Softmax activations}
\begin{figure}
	\centering
        \subfloat[DeiT-S]{
	  \includegraphics[width=0.92\linewidth]{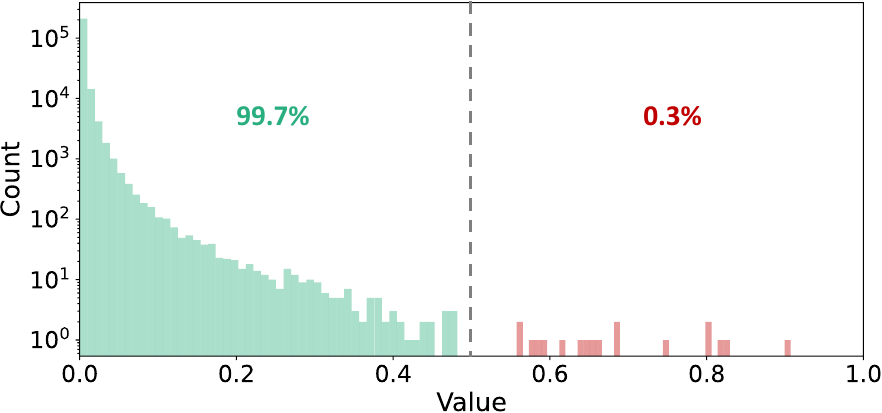}
	} \\
        \subfloat[LLaMA-7B]{
	  \includegraphics[width=0.92\linewidth]{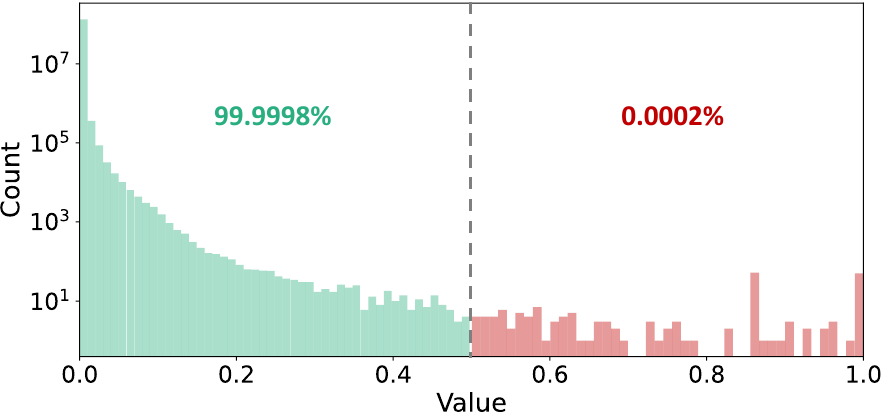}
	}
        
	\caption{Histograms of the first module’s Softmax activations in DeiT-S and LLaMA-7B. It can be clearly seen that the distributions are extremely unbalanced, with the vast majority concentrated on small values and a few dispersed on large values.}
	\label{fig:softmax}
\end{figure}

We conduct a statistical examination of the numerical distributions of Softmax activations, and the histograms of the DeiT-S and LLaMA-7B models are illustrated in Fig. \ref{fig:softmax}. It is evident that these activation values predominantly adhere to a power-law distribution, deviating from the common Gaussian or Laplace distributions. This results in a highly unbalanced distribution pattern. Specifically, in the case of the LLaMA-7B model, an overwhelming majority (99.9998\%) of the values are concentrated within the (0, 0.5] interval, and conversely, a minuscule fraction (0.0002\%) of values is dispersed across the (0.5, 1.0] interval.
Notably, the values falling within the (0.5, 1.0] interval carry significant implications, particularly in representing critical relationships between different tokens in the self-attention mechanism. Despite their relatively small proportion, these values embody essential information and, as such, should not be dismissed as mere outliers and naively clipped. Instead, it is imperative to carefully preserve these values to maintain the capacity of attention.

Such unbalanced distributions present significant obstacles for traditional quantization methods, rendering both uniform and log2 quantization poorly accommodated. The former sets quantization intervals uniformly throughout the entire range, leading to excessively few levels in the centralized (0, 0.5] interval, and the latter is overly dispersed in describing values within the (0.5, 1.0] interval.
In conclusion, it is crucial to develop an innovative quantizer that effectively represents values across both intervals simultaneously.

\subsection{Scale Reparameterization}
Thanks to the quantization-inference decoupling paradigm, we can first employ complex quantizers to accommodate the extreme distributions of the above activations, and then we convert them to simplified quantizers for inference via mathematically equivalent transformations, which ensures both quantization accuracy and inference efficiency. 
Here, the equivalent transformation is realized by our proposed scale reparameterization, and we detail the scale reparameterization methods for LayerNorm activations and Softmax activations, respectively, in the following.

\subsubsection{Scale reparameterization for LayerNorm activations}
\begin{figure}
	\centering
	\includegraphics[width=1.0\linewidth]{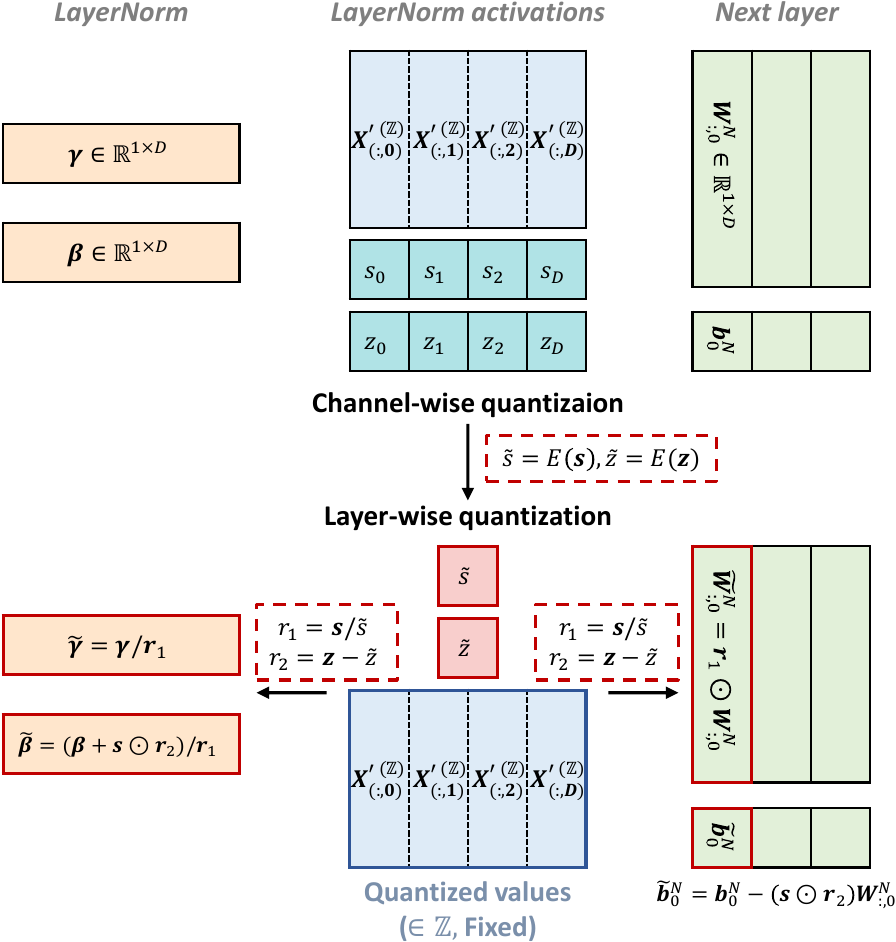}
	\caption{Illustration of the scale reparameterization method for LayerNorm activations. Initially, we apply channel-wise quantization. Then, we convert it to layer-wise quantization by adjusting LayerNorm's affine factors and the next layer's parameters through equivalent transformations. Note that the quantized integer weights are fixed in this process.}
	\label{fig:layernorm_rep}
\end{figure}

For LayerNorm activations, we apply channel-wise quantization in the quantization process and layer-wise quantization in the inference process. We will introduce the scale reparameterization method that transforms channel-wise quantization equivalently to layer-wise quantization.
To begin with, LayerNorm is utilized to normalize the input along the hidden feature dimension, and the calculation process is formulated as follows:
\begin{equation}
\bm{X}'_{n,:} = \text{LayerNorm}(\bm{X}_{n,:}) = \frac{\bm{X}_{n,:}-\text{E}[\bm{X}_{n,:}]}{\sqrt{\text{Var}[\bm{X}_{n,:}]+\epsilon}}\odot \bm{\gamma} + \bm{\beta},
\end{equation}
where $\bm{X},\bm{X}' \in \mathbb{R}^{N\times D}$ \footnote{To simplify the formulation, we ignore the batch dimension.} are the input and activation of LayerNorm, respectively, $N$ is the number of embedded tokens, $D$ is the size of hidden features, and $n=1,2,\cdots,N$. Here, $\text{E}[\bm{X}_{n,:}]$ and $\text{Var}[\bm{X}_{n,:}]$ are the mean and variance, respectively, and $\bm{\gamma}\in \mathbb{R}^{1\times D}$ and $\bm{\beta}\in \mathbb{R}^{1\times D}$ are the row vectors of linear affine factors. $\odot$ denotes Hadamard product.

Given the LayerNorm activation $\bm{X}'$, we initially apply channel-wise quantization and obtain the quantization scale $\bm{s}\in \mathbb{R}^{1\times D}$ and zero-point $\bm{z}\in \mathbb{Z}^{1\times D}$.
Our objective is to transform them into a uniform format: $\tilde{\bm{s}}=\tilde{s} \cdot \bm{1}$ and $\tilde{\bm{z}}=\tilde{z} \cdot \bm{1}$, where $\bm{1}$ denotes a $D$-dimensional row vector consisting entirely of ones, and the scalar quantities $\tilde{s}\in \mathbb{R}^{1}$ and $\tilde{z}\in \mathbb{Z}^{1}$ are prepared for layer-wise quantization. In this paper, we designate $\tilde{s}$ and $\tilde{z}$  as the respective mean values of original $\bm{s}$ and $\bm{z}$, i.e., $\tilde{s}=\text{E}[\bm{s}], \tilde{z}=\text{E}[\bm{z}]$. Then we define two variation factors: $\bm{r}_1=\bm{s}/\tilde{\bm{s}}$ \footnote{Here, the vector division is an element-wise operation like Hadamard product.} and $\bm{r}_2=\bm{z}-\tilde{\bm{z}}$, and the following equations are valid: 
\begin{align}
  \label{eq:3.2-1} \tilde{\bm{z}} &= \bm{z}-\bm{r}_2 = \left\lfloor -\frac{\left[\min(\bm{X}'_{:,d})\right]_{1\leq d \leq D}+\bm{s}\odot \bm{r}_2}{\bm{s}} \right\rceil, \\
  \label{eq:3.2-2} \tilde{\bm{s}} &= \frac{\bm{s}}{\bm{r}_1} = \frac{\left[\max(\bm{X}'_{:,d})-\min(\bm{X}'_{:,d})\right]_{1\leq d \leq D}/\bm{r}_1}{2^b-1}.
\end{align}

Eq. \ref{eq:3.2-1} illustrates that we can obtain $\tilde{\bm{z}}$ by adding $\bm{s}\odot \bm{r}_2$ to each channel of $\bm{X}'$, and Eq. \ref{eq:3.2-2} demonstrates that dividing each channel of $\bm{X}'$ by $\bm{r}_1$ results in $\tilde{\bm{s}}$. Conveniently, these transformations can be effectively implemented by adjusting the LayerNorm’s affine factors as follows:
\begin{equation}
    \label{eq:raparm_1}
  \widetilde{\bm{\beta}} = \frac{\bm{\beta}+\bm{s}\odot \bm{r}_2}{\bm{r}_1}, \quad \widetilde{\bm{\gamma}} = \frac{\bm{\gamma}}{\bm{r}_1}.
\end{equation}

The outlined procedure successfully achieves the reparameterization of $\tilde{\bm{s}}$ and $\tilde{\bm{z}}$, while this leads to a distribution shift of activations, i.e., $\widetilde{\bm{X}}'_{n,:}=(\bm{X}'_{n,:}+\bm{s}\odot \bm{r}_2)/\bm{r}_1$. Fortunately, this shift can be nullified through the inverse compensation applied to the next layer's weights and biases. Specifically, by employing equivalent transformations, it is demonstrated that:
\begin{equation}
\begin{split}
  \bm{X}'_{n,:}\bm{W}^{N}_{:,j}+\bm{b}^{N}_j = \frac{\bm{X}'_{n,:}+\bm{s}\odot \bm{r}_2}{\bm{r}_1} \left(\bm{r}_1\odot\bm{W}^{N}_{:,j}\right) \\ + \left(\bm{b}^{N}_j - (\bm{s}\odot \bm{r}_2) \bm{W}^{N}_{:,j}\right),
\end{split}
\end{equation}
where $j=1,2,\cdots,D_h$.
Therefore, to maintain consistency of the next layer's outputs before and after reparameterization, the weights and biases need to be modified accordingly:
\begin{equation}
\label{eq:raparm_2}
\begin{aligned}
  \widetilde{\bm{W}}^{N}_{:,j} &= \bm{r}_1\odot\bm{W}^{N}_{:,j}, \\
  \widetilde{\bm{b}}^{N}_j &= \bm{b}^{N}_j - (\bm{s}\odot \bm{r}_2) \bm{W}^{N}_{:,j}.    
\end{aligned}
\end{equation}


At this point, through the equivalent and interpretable adjustments of the LayerNorm’s affine factors, $\widetilde{\bm{\beta}}$ and $\widetilde{\bm{\gamma}}$, alongside the next layer’s weights and biases, $\widetilde{\bm{W}}^{N}$ and $\widetilde{\bm{b}}^{N}$, we methodically reparameterize channel-wise quantization ($\bm{s}$ and $\bm{z}$) to layer-wise quantization ($\tilde{s}$ and $\tilde{z}$). 
Fig. \ref{fig:layernorm_rep} provides a visual representation of the reparameterization process.
Note that in this process, the quantized integer values of LayerNorm activations are fixed, which on one hand optimizes the computational effort, and on the other hand maintains the current state in the quantization space undisturbed, thus potentially keeping the minimized quantization bias.


\subsubsection{Scale reparameterization for Softmax activations}
\begin{figure}
	\centering
        \subfloat[Quant]{
	  \includegraphics[width=0.98\linewidth]{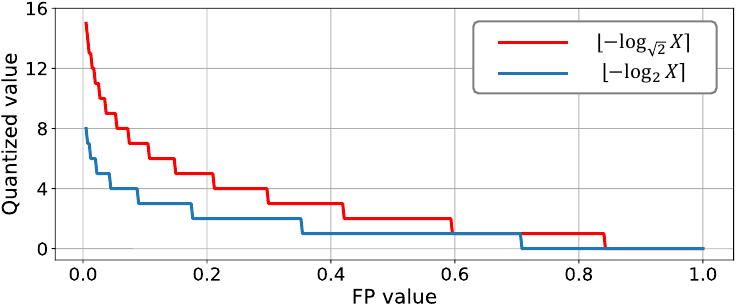}
        \label{fig:log_quant_1}
	} \\
        \subfloat[DeQuant]{
	  \includegraphics[width=0.98\linewidth]{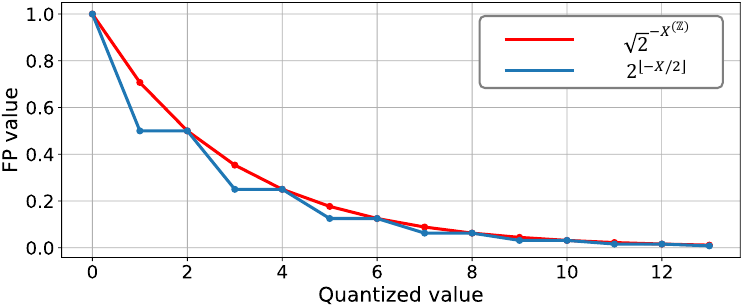}
        \label{fig:log_quant_2}
	}
        
	\caption{Comparison of the log2 and log$\sqrt{2}$ quantizers. In the quantization process, the log$\sqrt{2}$ quantizer can better represent the power-law distribution; and in the dequantization process, it can be equivalently transformed to the log2 quantizer by adjusting the quantization scale.}
	\label{fig:log_quant}
\end{figure}

Compared to the uniform and log2 quantizers, the log$\sqrt{2}$ quantizer offers superior expressiveness for Softmax activations.  This comparison, particularly between the log2 and log$\sqrt{2}$ quantizers, is depicted in Fig. \ref{fig:log_quant_1}. The log$\sqrt{2}$ quantizer stands out by providing a denser and more nuanced representation for larger values, unlike the log2 quantizer, which tends to offer a sparser description. For example, when considering $s$ = 1, the log2 quantizer rounds all values in the interval [$2^{-1.5}$, $2^{-0.5}$], i.e., [0.35, 0.71], indiscriminately to $2^{-1}$, leading to a significant loss of information. In contrast, the log$\sqrt{2}$ quantizer rounds values in the narrower interval [$\sqrt{2}^{-1.5}$, $\sqrt{2}^{-0.5}$], i.e., [0.59, 0.84], to $\sqrt{2}^{-1}$, effectively mitigating the above issue.
Importantly, the log$\sqrt{2}$ quantizer can be equivalently transformed into the log2 quantizer, and can also benefit from efficient bit-shifting operations during inference.

In the following, we will introduce in detail the scale reparameterization methods that change the base in both quantization and de-quantization procedures.
Specifically, given the Softmax activation $\bm{A}$ and the scale $s\in \mathbb{R}^1$ of log$\sqrt{2}$ quantizer, by employing the base changing formula of the logarithmic function, we can derive:
\begin{equation}
\label{eq:3.3-1}
\begin{split}
    \bm{A}^{(\mathbb{Z})} &= \text{clip}\left(\left\lfloor -\log_{\sqrt{2}} \frac{\bm{A}}{s} \right\rceil, 0, 2^b-1 \right) \\ 
    &= \text{clip}\left(\left\lfloor -2\log_2 \frac{\bm{A}}{s} \right\rceil , 0, 2^b-1 \right).
\end{split}
\end{equation}

Eq. \ref{eq:3.3-1} shows that in the quantization procedure, conversion to the log2 quantizer is straightforward, necessitating only multiplication by a constant factor. Likewise, during de-quantization, we employ the base changing formula of the exponential function to convert to base-2. Nevertheless, the resulting exponential term $-\frac{\bm{A}^{(\mathbb{Z})}}{2}$ may not always yield an integer, which is crucial for conducting bit-shifting operations. 
Consequently, we examine the parity of $-\frac{\bm{A}^{(\mathbb{Z})}}{2}$ on a case-by-case basis as follows:
\begin{equation}
\begin{split}
    \widehat{\bm{A}} &= s\cdot \sqrt{2}^{-\bm{A}^{(\mathbb{Z})}} = s\cdot 2^{-\frac{\bm{A}^{(\mathbb{Z})}}{2}} \\
    &= \begin{cases}
        s\cdot 2^{-\frac{\bm{A}^{(\mathbb{Z})}}{2}} &\;\; \bm{A}^{(\mathbb{Z})}=2k, k\in \mathbb{Z} \\
        s\cdot 2^{-\frac{\bm{A}^{(\mathbb{Z})+1}}{2}}\cdot \sqrt{2} &\;\; \bm{A}^{(\mathbb{Z})}=2k+1, k\in \mathbb{Z}
    \end{cases} \\
    &= s\cdot 2^{\left\lfloor-\frac{\bm{A}^{(\mathbb{Z})}}{2}\right\rfloor}
    \cdot \left[\mathds{1}(\bm{A}^{(\mathbb{Z})})\cdot(\sqrt{2}-1)+1 \right],
\end{split}
\end{equation}
where $\left\lfloor\cdot\right\rfloor$ represents the floor function, ensuring that $\left\lfloor-\frac{\bm{A}^{(\mathbb{Z})}}{2}\right\rfloor$ is an integer, as shown in Fig. \ref{fig:log_quant_2}, and $\mathds{1}(\cdot)$ denotes a parity indicator function, which returns 0 for even numbers and 1 for odd numbers.

The parity indicator function and its associated coefficients can be integrated into $s$ to derive the reparameterized scale, denoted as $\tilde{s}$, in the following manner:
\begin{equation}
\label{eq:3.3-2}
    \tilde{s} = s\cdot \left[\mathds{1}(\bm{A}^{(\mathbb{Z})})\cdot(\sqrt{2}-1)+1 \right].
\end{equation}

By reparameterizing the scale to $\tilde{s}$, the de-quantization procedure gains the advantage of efficient bit-shifting operations. While this reparameterized scale $\tilde{s}$ adds minimal extra computational overhead during inference compared to the original scale $s$, it's important to note its efficiency: the parity indicator function can be computed swiftly, for instance, by checking the least significant bit of $\bm{A}^{(\mathbb{Z})}$ on FPGAs.

\subsection{Per-Channel Dual Clipping}
\begin{figure*}
	\centering
	\includegraphics[width=0.95\linewidth]{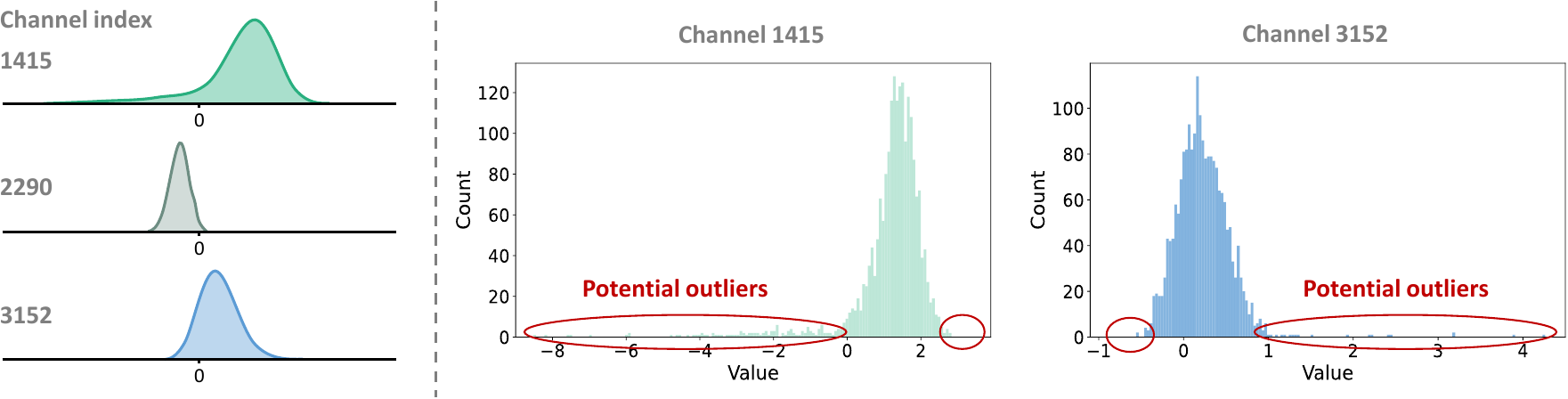}
	\caption{Histograms of different channels of the first module’s LayerNorm activations in LLaMA-7B. We can observe two features: i) the distribution within one channel is highly unbalanced, with extremely asymmetric outliers in the upper and lower bounds; and ii) the form and degree of asymmetry varies among channels, with left-biased patterns (channel 3152) as well as right-biased patterns (channel 1415). The above features highlight the significance of fine-grained outlier clipping methods to ensure accurate quantization.}
	\label{fig:dual-clip-1}
\end{figure*}

\begin{figure*}
	\centering
	\includegraphics[width=0.95\linewidth]{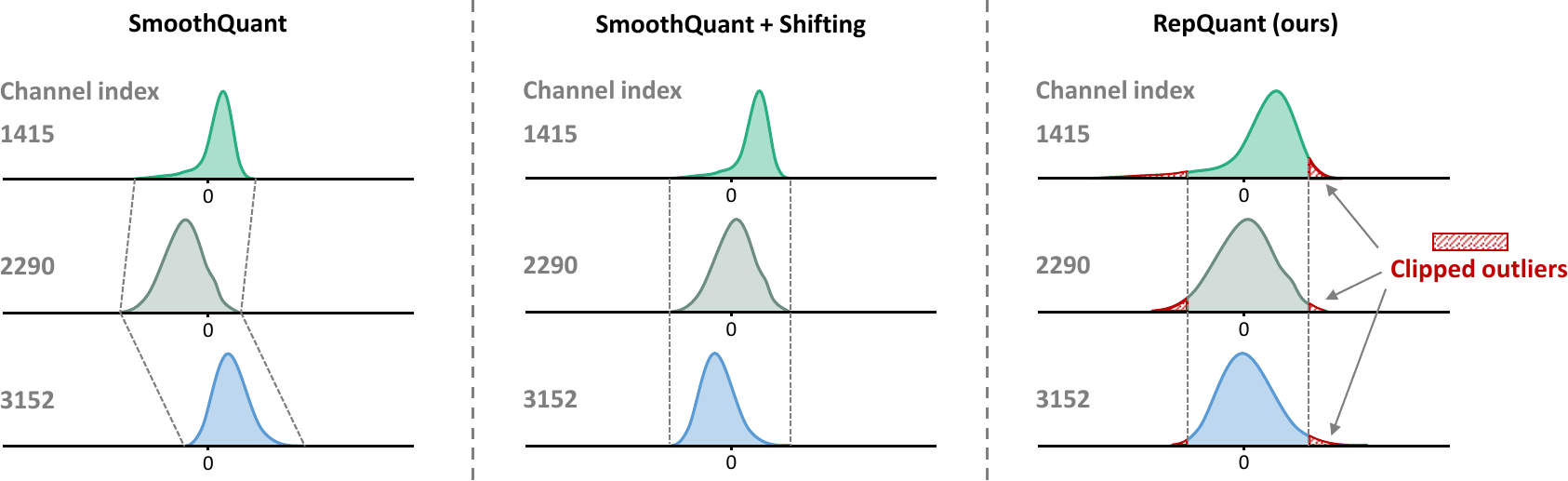}
	\caption{Comparison of the handling of LayerNorm activations between SmoothQuant and our proposed RepQuant. SmoothQuant simply aligns the distribution of each channel by scaling (or plus shifting) \emph{before} quantization. However, this alignment is implemented in the floating-point space, which cannot directly reflect the  quantization performance. Instead, RepQuant first performs channel-wise quantization, which in particular incorporates fine-grained outlier clipping, and then aligns \emph{after} quantization via scale reparameterization, thus promising to minimize bias in the actual quantization space.}
	\label{fig:dual-clip-2}
\end{figure*}

As mentioned above, we initially apply channel-wise quantization for LayerNorm activations. Unfortunately, due to the presence of systematic outliers, simply using extreme values as bounds adversely affects performance. Consequently, it's crucial to investigate effective methods for accurately clipping outliers.
To this end, we further conduct an in-depth analysis of the distribution of LayerNorm activations, and the histograms for different channels are illustrated in Fig. \ref{fig:dual-clip-1}. 

Our observations reveal two distinct characteristics: \underline{\textbf{i)}} the distribution within one channel is markedly unbalanced, characterized by pronounced asymmetric outliers at both the upper and lower extremes; and \underline{\textbf{ii)}} there is considerable variation in the form and extent of asymmetry across channels, exhibiting both left-biased (e.g., channel 3152) and right-biased patterns (e.g., channel 1415).
These insights collectively suggest two specific requirements for the clipping method: \underline{\textbf{i)}} each channel should employ different clipping coefficients for the upper and lower bounds to deal with the asymmetry; and \underline{\textbf{ii)}} different channels necessitates different pairs of clipping coefficients to accommodate the varying forms of asymmetry.
Thus, we are motivated to develop a per-channel dual clipping method, which precisely meets the above requirements.

A select few studies have made preliminary attempts at dual clipping for activations, and one noteworthy scheme is a search-based approach~\cite{tu2023toward}. However, a non-negligible concern is the computational overhead. It determines the upper and lower bounds iteratively through tedious loops, which is potentially feasible for layer-wise quantization where only two scalar values are identified. However, for channel-wise quantization, this approach necessitates a comprehensive search for each channel, becoming impractical for architectures with thousands of channels (e.g., LLaMA-7B).

To address the above issues, we propose a learning-based approach that can efficiently identify outliers for each channel at a fine-grained level. Specifically, we define two learnable clipping factors: $\bm{\alpha}_1, \bm{\alpha}_2\in \mathbb{R}^{1\times D}$, which, after a Sigmoid function, produce a contraction of [0, 1] on the extremes of each channel, respectively, and thus yield the initial upper and lower bounds for calibration in Eq. \ref{eq:sz} as follows:

\begin{align}
\label{eq:clip}
\bm{X}'_{up} = \left[\max(\bm{X}'_{:,d})\right]_{1\leq d \leq D} \odot \text{Sigmoid}(\bm{\alpha}_1), \\
\label{eq:clip-1}
\bm{X}'_{low} = \left[\min(\bm{X}'_{:,d})\right]_{1\leq d \leq D} \odot \text{Sigmoid}(\bm{\alpha}_2).
\end{align}

Here, it is demonstrated that compared to direct numerical contraction, the utilization of the Sigmoid function provides better and more stable optimization (please see Table \ref{tab:ablation_4} for details).
Then, we learn the optimal clipping factors by minimizing the discrepancy in the quantization space with the following optimization objective:
\begin{equation}
\label{eq:clip-opt}
\left(\bm{\alpha}^*_1, \bm{\alpha}^*_2\right) = \mathop{\arg\min}_{\bm{\alpha}_1, \bm{\alpha}_2} ||\bm{X}'-\widehat{\bm{X}}'||^2 .
\end{equation}

Note that this optimization process is very efficient and can typically be completed in a few seconds for one layer on a single NVIDIA A6000 GPU.

In addition, we further emphasize the advantages of RepQuant, which surpasses other activation-oriented quantization methods, particularly SmoothQuant, through its integration of per-channel dual clipping. A detailed comparison is showcased in Fig. \ref{fig:dual-clip-2}. Unlike SmoothQuant, which merely adjusts the distribution of each channel through scaling (or plus shifting) in the floating-point space \emph{before} quantization, RepQuant adopts a more effective approach. It begins with channel-wise quantization, notably featuring fine-grained outlier clipping, and then proceeds to align the distribution using scale reparameterization \emph{after} quantization. This ensures a more accurate reflection of quantization performance, significantly reducing bias in the actual quantization space.

\subsection{Sequential Quantization Pipeline}

\begin{algorithm}[t]
	\caption{Sequential Pipeline of RepQuant}
	\label{alg:RepQuant}
	\KwIn{Pre-trained full-precision transformer model,}

        \qquad \quad  Calibration dataset.

        \For{all i = 1, 2, $\cdots$, L-th layer in pre-trained model}{
        Collect the $i$-th layer's input data $\bm{x}^{(i)}$;

        \textcolor[RGB]{40,125,142}{\# Activation quantization}

        \If{$\bm{x}^{(i)}$ is LayerNorm activation}{
        {\color[RGB]{218,128,128}\quad \textbf{channel-wise quant $\to$ layer-wise quant}}
        
        Initialize per-channel bounds $\bm{X}'_{up}$ and $\bm{X}'_{low}$ by Eqs. \ref{eq:clip} and \ref{eq:clip-1}; 
        
        Initialize channel-wise quant ($\bm{s}$,$\bm{z}$) for $\bm{x}^{(i)}$;

        \textcolor[RGB]{40,125,142}{\# Per-channel dual clipping}
        
        Learn clipping factors $\bm{\alpha}_1$ and $\bm{\alpha}_2$ by Eq. \ref{eq:clip-opt};

        \textcolor[RGB]{40,125,142}{\# Scale reparameterization}

        Update the quantizer $\tilde{s}=\text{E}[\bm{s}]$ and $\tilde{z}=\text{E}[\bm{z}]$;
          
        Calculate $\bm{r}_1=\bm{s}/(\tilde{s}\cdot\bm{1})$ and $\bm{r}_2=\bm{z}-(\tilde{z}\cdot\bm{1})$;
          
        Update LayerNorm’s affine factors $\widetilde{\bm{\beta}}$ and $\widetilde{\bm{\gamma}}$ by Eq. \ref{eq:raparm_1};

        Update $i$-th layer's weights $\widetilde{\bm{W}}^{N}$ and biases $\widetilde{\bm{b}}^{N}$ by Eq. \ref{eq:raparm_2};


        }
        
        \ElseIf{$\bm{x}^{(i)}$ is Softmax activation}{
        {\color[RGB]{218,128,128}\qquad \qquad \textbf{log2 quant} $\to$ \textbf{log$\sqrt{2}$ quant}}

        Initialize log$\sqrt{2}$ quant ($s$) for $\bm{x}^{(i)}$;

        \textcolor[RGB]{40,125,142}{\# Scale reparameterization}

        Update quant procedure by Eq. \ref{eq:3.3-1};

        Update $\tilde{s}$ in de-quant procedure by Eq. \ref{eq:3.3-2};
        }

        \Else{
        Perform uniform quant for $\bm{x}^{(i)}$;

        }
        \textcolor[RGB]{40,125,142}{\# Weight quantization}
        
        Perform GPTQ to quantize $i$-th layer's weights;
        
        }

        \KwOut{Quantized transformer model.}

\end{algorithm}

Although RepQuant focuses on the challenges of activation quantization, the entire framework provides excellent compatibility for off-the-shelf weight quantization methods to comprehensively ensure quantization performance. In this paper, we adopt GPTQ~\cite{frantar2022gptq}, an efficient weight reconstruction method designed for large transformer models. Importantly, GPTQ can be seamlessly integrated, resulting in a sequential quantization pipeline in a layer-by-layer fashion. In particular, for LayerNorm activations, during the scale reparameterization process from channel-wise quantization to layer-wise quantization, the floating-point weights of the next layer are updated. In this way, GPTQ can be performed directly on the updated weights, which eliminates the re-calibration procedure in the previous version RepQ-ViT. Consequently, to obtain the final quantized model, it is sufficient for RepQuant to perform a one-time sequential traversal of the inputs and weights of each layer, without additional post-processing. The whole sequential pipeline is summarized in Algorithm \ref{alg:RepQuant}.

\section{Experiments}
\subsection{Experimental Setup}
\label{sec:setup}
\textbf{Models and datasets} We evaluate RepQuant on different transformer variants on multiple tasks, including vision, language, and multi-modal transformers. The adopted models and datasets are described below.
\begin{itemize}
\item Vision transformers: For image classification, we evaluate the quantization performance on ViT~\cite{dosovitskiy2020image}, DeiT~\cite{touvron2021training}, and Swin~\cite{liu2021swin} on ImageNet \cite{deng2009imagenet} dataset. The pre-trained models are obtained from timm\footnote{\url{https://github.com/rwightman/pytorch-image-models}}. For object detection, we adopt Mask R-CNN~\cite{he2017mask} and Cascade Mask R-CNN~\cite{cai2018cascade} with Swin as the backbone and perform evaluation on COCO~\cite{lin2014microsoft} dataset using mmdetection\footnote{\url{https://github.com/open-mmlab/mmdetection}}.
In addition, we also quantize SAM~\cite{kirillov2023segment} and evaluate its performance of zero-shot instance segmentation\footnote{\url{https://github.com/RockeyCoss/Prompt-Segment-Anything}} on COCO dataset.
\item Language transformers: We adopt two popular large models: OPT~\cite{zhang2022opt} and LLaMA~\cite{touvron2023llama} with various scales. The pre-trained models are obtained from huggingface\footnote{\url{https://huggingface.co/models}}. We report the perplexity on multiple datasets including WikiText2~\cite{merity2016pointer}, Pen Treebank~\cite{marcus1994penn}, and C4~\cite{raffel2020exploring}, and also evaluate the accuracy of zero-shot tasks including PIQA~\cite{bisk2020piqa}, ARC~\cite{clark2018think}, HellaSwag~\cite{zellers2019hellaswag}, and Winogrande~\cite{sakaguchi2021winogrande}.
\item Multi-modal transformers: CLIP~\cite{radford2021learning}, which learns the ability of image-text matching, is served as the baseline model. We quantize both the image encoder and text encoder and evaluate the zero-shot image classification performance on CIFAR100 and ImageNet datasets.
\end{itemize}

\textbf{Implementation details}
All experiments in this paper are implemented in Pytorch.
For vision transformers, the calibration datasets contain 1024 random samples, which is aligned with GPTQ for small models~\cite{frantar2022optimal}, and for language transformers, we use 128 random samples for calibration following standard GPTQ~\cite{frantar2022gptq}.
We quantize the weights and activations of all components in transformers to low bits, including LayerNorm activations and Softmax activations.
In inference, weights are applied channel-wise quantization and activations are applied layer-wise quantization, which guarantees hardware efficiency. All of them are employed uniform quantizers except Softmax activations which are employed log quantizers.
The default parameter calibration strategy is Percentile, while we apply the proposed fine-grained dual clipping for LayerNorm activations.
When solving Eq. \ref{eq:clip-opt} in dual clipping, we adopt Adam~\cite{kingma2014adam} optimizer with only 100 iterations and a learning rate of 0.01.
All experiments are conducted on NVIDIA A6000 GPUs.

\textbf{Comparison methods}
For vision transformers, the previous version RepQ-ViT~\cite{li2023repq} served as the strong baseline, and its results are directly accessed from the original paper. For language Transformers, our RepQuant is exhaustively compared to influential works, including SmoothQuant~\cite{xiao2023smoothquant}, Outlier Suppression+~\cite{wei2023outlier}, RPTQ~\cite{yuan2023rptq}, and OmniQuant~\cite{shao2023omniquant}. However, it would be unfair to conduct comparisons directly due to misalignment of quantization settings, e.g., in their default configurations, RPTQ consistently retains LayerNorm activations and Softmax activations as 8bit, and OmniQuant fixes Softmax activations to 16bit and applies per-token dynamic quantization to all activations. Therefore, we need to reproduce these works with our settings to ensure fair comparisons. Also, we find that low-bit uniform quantization for Softmax activations leads to crashes, so we still reserve them as 8-bit. To summarize, we reproduce them with the following setup: quantize all activations to low-bit except Softmax activations, and apply layer-wise quantization to all activations.

\subsection{Performance Evaluation on Vision Transformers}
We adequately demonstrate the effectiveness of the proposed RepQuant by quantizing vision transformers on the image classification and object detection tasks. Furthermore, we also evaluate the zero-shot instance segmentation performance of the quantized SAM models. In the following, we will present and discuss the experimental results in turn.

\begin{table*}[t]
\centering
\setlength{\tabcolsep}{9pt}
\small
\caption{Quantization results of image classification on ImageNet dataset, with each data representing the Top-1 accuracy (\%) of the quantized model. Here, ``Prec. (W/A)'' denotes the quantization precision, i.e., the weights and activations are quantized to W and A bits, respectively.}
\begin{tabular}{ccccccccc}
\toprule
\textbf{Method} &\textbf{Prec. (W/A)} & \textbf{ViT-S }& \textbf{ViT-B} & \textbf{DeiT-T }& \textbf{DeiT-S} & \textbf{DeiT-B} & \textbf{Swin-S }& \textbf{Swin-B} \\
\midrule
Full-Precision & 32/32 & 81.39 & 84.54 & 72.21 & 79.85 & 81.80 & 83.23 & 85.27 \\
\midrule
FQ-ViT \cite{lin2021fq} & 4/4 & 0.10 & 0.10 & 0.10 & 0.10 & 0.10 & 0.10 & 0.10 \\
PTQ4ViT \cite{yuan2021ptq4vit} & 4/4 & 42.57 & 30.69 & 36.96 & 34.08 & 64.39 & 76.09 & 74.02 \\
APQ-ViT \cite{ding2022towards} & 4/4 & 47.95 & 41.41 & 47.94 & 43.55 & 67.48 & 77.15 & 76.48 \\
RepQ-ViT \cite{li2023repq} & 4/4 & 65.05 & 68.48 & 57.43 & 69.03 & 75.61 & 79.45 & 78.32 \\
RepQuant (ours) & 4/4 & \textbf{73.28} & \textbf{77.84} & \textbf{64.44} & \textbf{75.21} & \textbf{78.46} & \textbf{81.52} & \textbf{82.80} \\
\midrule
FQ-ViT \cite{lin2021fq} & 6/6 & 4.26 & 0.10 & 58.66 & 45.51 & 64.63 & 66.50 & 52.09 \\
PSAQ-ViT \cite{li2022patch} & 6/6 & 37.19 & 41.52 & 57.58 & 63.61 & 67.95 & 72.86 & 76.44 \\
Ranking \cite{liu2021post} & 6/6 & - & 75.26 & - & 74.58 & 77.02 & - & - \\
PTQ4ViT \cite{yuan2021ptq4vit} & 6/6 & 78.63 & 81.65 & 69.68 & 76.28 & 80.25 & 82.38 & 84.01 \\
APQ-ViT \cite{ding2022towards} & 6/6 & 79.10 & 82.21 & 70.49 & 77.76 & 80.42 & 82.67 & 84.18 \\
NoisyQuant\cite{liu2023noisyquant} & 6/6 & 78.65 & 82.32 & - & 77.43 & 80.70 & 82.86 & 84.68 \\
RepQ-ViT \cite{li2023repq} & 6/6 & 80.43 & 83.62 & 70.76 & 78.90 & 81.27 & 82.79 & 84.57 \\
RepQuant (ours) & 6/6 & \textbf{80.51} & \textbf{83.75} & \textbf{70.89} & \textbf{79.06} & \textbf{81.41} & \textbf{82.93} & \textbf{84.86} \\
\bottomrule
\end{tabular}

\label{tab:vit-imagenet}
\end{table*}

\begin{table*}[t]
\centering
\small
\setlength{\tabcolsep}{8pt}
\caption{Quantization results of object detection and instance segmentation on COCO dataset. ``AP$^\text{box}$'' is the box average precision for object detection, and ``AP$^\text{mask}$'' is the mask average precision for instance segmentation. Here, ``Prec. (W/A)'' denotes the quantization precision, i.e., the weights and activations are quantized to W and A bits, respectively.}
\label{tab:coco}
\begin{tabular}{cccccccccc}
\toprule
\multirow{3.5}{*}{\textbf{Method}} & \multirow{3.5}{*}{\textbf{Prec. (W/A)}} & \multicolumn{4}{c}{\textbf{Mask R-CNN}} & \multicolumn{4}{c}{\textbf{Cascade Mask R-CNN}} \\
\cmidrule(lr){3-6}\cmidrule(lr){7-10}
&& \multicolumn{2}{c}{\textbf{w. Swin-T}} & \multicolumn{2}{c}{\textbf{w. Swin-S}} & \multicolumn{2}{c}{\textbf{w. Swin-T}} & \multicolumn{2}{c}{\textbf{w. Swin-S}} \\
&& \textbf{AP}$^\text{\textbf{box}}$ & \textbf{AP}$^\text{\textbf{mask}}$ & \textbf{AP}$^\text{\textbf{box}}$ & \textbf{AP}$^\text{\textbf{mask}}$ & \textbf{AP}$^\text{\textbf{box}}$ & \textbf{AP}$^\text{\textbf{mask}}$ & \textbf{AP}$^\text{\textbf{box}}$ & \textbf{AP}$^\text{\textbf{mask}}$ \\
\midrule
Full-Precision & 32/32 & 46.0 & 41.6 & 48.5 & 43.3 & 50.4 & 43.7 & 51.9 & 45.0 \\
\midrule
PTQ4ViT \cite{yuan2021ptq4vit} & 4/4 & 6.9 & 7.0 & 26.7 & 26.6 & 14.7 & 13.5 & 0.5 & 0.5  \\
APQ-ViT \cite{ding2022towards} & 4/4 & 23.7 & 22.6 & \textbf{44.7} & 40.1 & 27.2 & 24.4 & 47.7 & 41.1 \\
RepQ-ViT \cite{li2023repq} & 4/4 & 36.1 & 36.0 & 44.2 & 40.2 & 47.0 & 41.4 & 49.3 & 43.1 \\
RepQuant (ours) & 4/4 & \textbf{37.2} & \textbf{36.8} & 44.5 & \textbf{40.5}& \textbf{48.0} & \textbf{42.1} & \textbf{50.4} & \textbf{43.6} \\
\midrule
PTQ4ViT \cite{yuan2021ptq4vit} & 6/6 & 5.8 & 6.8 & 6.5 & 6.6 & 14.7 & 13.6 & 12.5 & 10.8 \\
APQ-ViT \cite{ding2022towards} & 6/6 & \textbf{45.4} & 41.2 & 47.9 & 42.9 & 48.6 & 42.5 & 50.5 & 43.9 \\
RepQ-ViT \cite{li2023repq} & 6/6 & 45.1 & 41.2 & 47.8 & 43.0 & 50.0 & 43.5 & 51.4 & 44.6 \\
RepQuant (ours) & 6/6 & 45.3 & \textbf{41.3} & \textbf{48.1} & \textbf{43.2} & \textbf{50.1} & \textbf{43.5} & \textbf{51.6} & \textbf{44.7} \\
\bottomrule

\end{tabular}
\end{table*}

\textbf{Quantization results for image classification}
We begin by discussing the quantization performance for image classification on ImageNet dataset, and the results are reported in Table~\ref{tab:vit-imagenet}. To highlight the advantages of RepQuant, our discussion centers on the low-bit cases, including W4/A4 and W6/A6 quantization. Notably, in contrast to existing methods that cause non-trivial performance degradation, the previous version RepQ-ViT is the first work to bring W4/A4 quantization to the usable level. Taking DeiT-S as an example, FQ-ViT crashes and only yields 0.10\% accuracy, and PTQ4ViT and APQ-ViT also underperform, experiencing accuracy declines of 45.77\% and 36.30\%, respectively. In contrast, RepQ-ViT markedly diminishes this accuracy drop to just 10.82\%. Although these improvements are considerable, a noticeable gap remains compared to the performance of the full-precision model. 
To this end, RepQuant, built on top of RepQ-ViT, introduces more advanced modules to further push the limits of quantization performance. For instance, when performing W4/A4 quantization, RepQuant presents encouraging 8.23\% and 9.36\% improvements over RepQ-ViT for ViT-S and ViT-B, respectively. Additionally, RepQuant consistently delivers substantial accuracy gains for various DeiT variants, with increments of 7.01\%, 6.18\%, and 2.85\%, respectively. Furthermore, in W6/A6 quantization, RepQuant attains an accuracy almost on par with the full-precision baseline, showing less than 1\% accuracy loss for most models.

\textbf{Quantization results for object detection}
The quantization results for object detection and instance segmentation tasks on COCO dataset are presented in Table~\ref{tab:coco}. We also focus on the performance of low-bit quantization. The complex model architectures in these tasks make previous quantization methods less effective. For example, PTQ4ViT becomes unfeasible across different bit-widths, while APQ-ViT suffers from notable accuracy loss in low-bit (e.g., W4/A4) cases and exhibits unstable performance for different backbones. RepQ-ViT outperforms these methods for different bit-width configurations and backbones, but the results are still far from satisfactory.
Fortunately, RepQuant consistently achieves more advanced performance than RepQ-ViT, showcasing high robustness.
In the case of W4/A4 quantization for Mask R-CNN framework with Swin-T, RepQuant improves over RepQ-ViT 1.1 box AP and 0.8 mask AP. Similarly, for Cascade Mask R-CNN framework, RepQuant enhances the box AP by 1.0 and mask AP by 0.7.
In addition, RepQuant achieves better results than RepQ-ViT on all models in W6/A6 quantification, with only a slight accuracy decrease compared to the full-precision baseline.
For instance, when quantizing Cascade Mask R-CNN framework with Swin-S, RepQuant reached 51.6 box AP and 44.7 mask AP, which is only 0.3 lower than the full-precision baseline for both box AP and mask AP. Similar results can also be obtained when Swin-T serves as the backbone, narrowing the box AP and mask AP gap with full-precision models to 0.3 and 0.2, respectively.

\begin{figure*}
	\centering
	\includegraphics[width=0.98\linewidth]{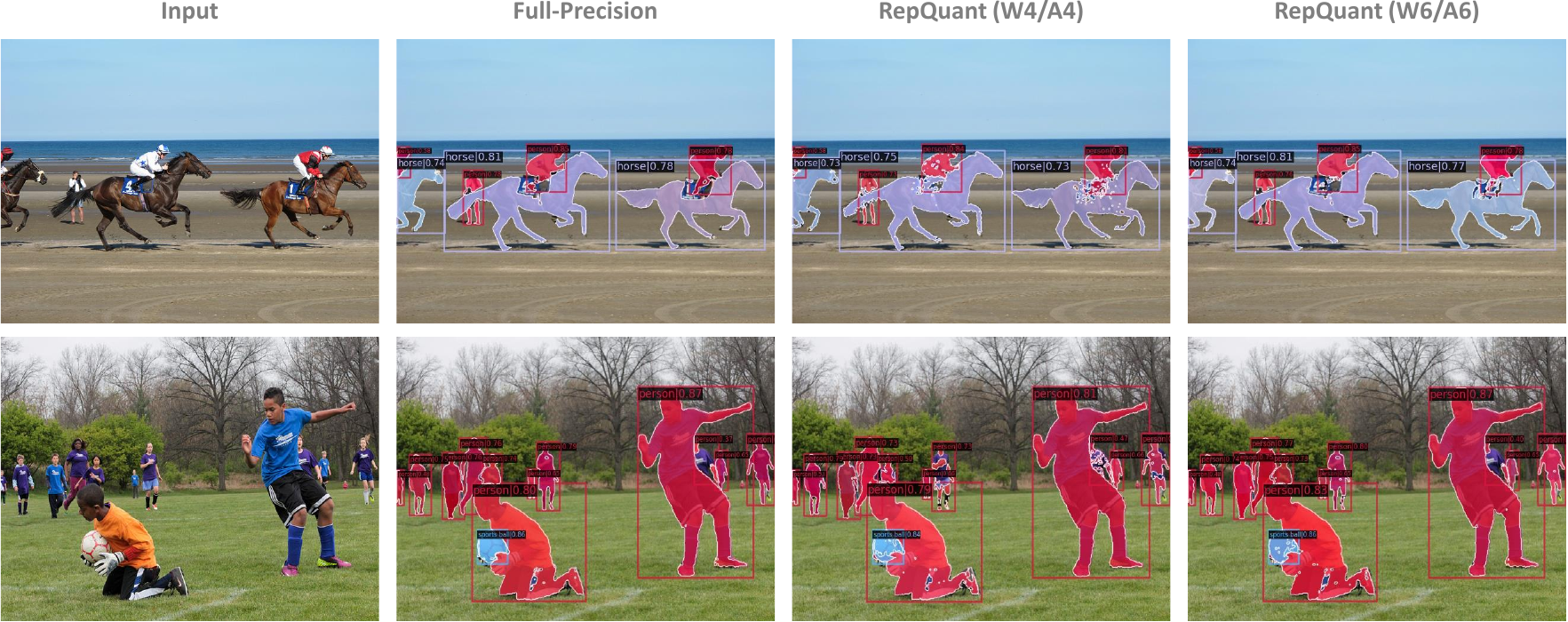}
	\caption{Visualization of zero-shot instance segmentation results of full-precision and quantized SAM models. Here, ViT-B serves as the image encoder, and it is prompted with H-Deformable-DETR w. Swin-L boxes. As one can see, RepQuant at W6/A6 precision can obtain the comparable results with the full-precision baseline. In W4/A4 case, the quantization performance is slightly degraded, but RepQuant still maintains the essential recognition ability, with minor shortcomings in segmentation wholeness and connectivity.}
	\label{fig:sam}
\end{figure*}

\begin{table}[t]
\centering
\small
\setlength{\tabcolsep}{8pt}
\caption{Quantization results of zero-shot instance segmentation of SAM models on COCO dataset, where SAM is prompted with H-Deformable-DETR w. Swin-L boxes. ``AP$^\text{mask}$'' is the mask average precision for instance segmentation. Here, ``Prec. (W/A)'' denotes the quantization precision, i.e., the weights and activations are quantized to W and A bits, respectively.}
\label{tab:sam}
\begin{tabular}{cccc}
\toprule
\textbf{SAM Model} &     \textbf{Method}          & \textbf{Prec. (W/A)} & \textbf{AP}$^\text{\textbf{mask}}$ \\ \midrule
\multirow{8}{*}{ViT-B}
&Full-Precision  & 32/32       &  42.5 \\ \cmidrule(l){2-4} 
&PTQ4ViT~\cite{yuan2021ptq4vit}         &  4/4        &  8.4 \\
&RepQ-ViT \cite{li2023repq}        &  4/4        &  37.6 \\
&RepQuant (ours) &  4/4        &  \textbf{39.2} \\ \cmidrule(l){2-4} 
&PTQ4ViT~\cite{yuan2021ptq4vit}         &  6/6        &  8.9 \\
&RepQ-ViT \cite{li2023repq}        &  6/6        &  42.0 \\
&RepQuant (ours) &  6/6        &  \textbf{42.2} \\ \midrule
\multirow{8}{*}{ViT-L}
&Full-Precision  & 32/32       &  46.3 \\ \cmidrule(l){2-4} 
&PTQ4ViT~\cite{yuan2021ptq4vit}         &  4/4        &  11.3 \\
&RepQ-ViT \cite{li2023repq}        &  4/4        &  40.7 \\
&RepQuant (ours) &  4/4        &  \textbf{42.0} \\ \cmidrule(l){2-4} 
&PTQ4ViT~\cite{yuan2021ptq4vit}         &  6/6        &  13.5 \\
&RepQ-ViT \cite{li2023repq}        &  6/6        &  45.8 \\
&RepQuant (ours) &  6/6        &  \textbf{46.1} \\ \bottomrule
\end{tabular}
\end{table}

\textbf{Quantization results for segment anything}
We extend the quantization process to the SAM models, specifically evaluating their performance in zero-shot instance segmentation. It's important to note that since the image encoder is responsible for the bulk of computational demand, our quantization efforts are focused solely on this component, while the prompt encoder and mask decoder are maintained as floating point operations. The numerical results are shown in Table \ref{tab:sam}, where RepQuant demonstrates consistently impressive performance. For instance, with ViT-B as the image encoder, RepQuant achieves significant improvements at W4/A4 precision, outperforming PTQ4ViT by a substantial 30.8 mask AP and surpassing RepQ-ViT by 1.6 mask AP. Moreover, in W6/A6 quantization, the performance of RepQuant is remarkably close to the full-precision baseline, showing a mere 0.3 mask AP discrepancy. Similar trends are observed when employing ViT-L as the image encoder, further underscoring the effectiveness across various configurations.
We have also undertaken a visual analysis of the zero-shot instance segmentation results, as illustrated in Fig. \ref{fig:sam}. Notably, with RepQuant operating at W6/A6 precision, the results are on par with those of the full-precision baseline. In the case of W4/A4 quantization, although there is a slight decline in performance, RepQuant commendably retains core recognition capabilities, which is evidenced by only minor deficiencies in segmentation completeness and connectivity.

\begin{table*}[t]
\centering
\small
\setlength{\tabcolsep}{7pt}
\caption{Quantization results of OPT models on language generation datasets, including WikiText2 (WIKI), Pen Treebank (PTB), and C4. We report the perplexity scores of each quantized model. Here, ``Prec. (W/A)'' denotes the quantization precision.} 
\begin{threeparttable}
\begin{tabular}{@{}ccccccccccc@{}}
\toprule
\multirow{2.5}{*}{\textbf{{Method}}} & \multirow{2.5}{*}{\textbf{Prec. (W/A)}} & \multicolumn{3}{c}{\textbf{OPT-6.7B}} & \multicolumn{3}{c}{\textbf{OPT-13B}} & \multicolumn{3}{c}{\textbf{OPT-30B}} \\ \cmidrule(l){3-11} 
 &  & \textbf{WIKI} & \textbf{PTB} & \textbf{C4} & \textbf{WIKI} & \textbf{PTB} & \textbf{C4}& \textbf{WIKI} & \textbf{PTB} & \textbf{C4}\\ \midrule
Full-Precision & 32/32 & 10.86 & 13.09 & 11.74 & 10.13 & 12.34 & 11.20 & 9.56 & 11.84 & 10.69 \\ \midrule
SmoothQuant\tnote{}*~\cite{xiao2023smoothquant} & 4/4 & 21586 & 15854 & 17037 & 14284 & 10229 & 8490 & 7249 & 8624 & 7603 \\
Outlier Suppression+*~\cite{wei2023outlier} & 4/4 & 16.82 & 17.69 & 17.12 & 15.93 & 19.46 &18.25  & 15.29 &  17.31 & 15.22 \\
RPTQ*~\cite{yuan2023rptq} & 4/4 & 19.29 & 24.89 & 20.35 & 15.74 & 22.51 & 16.63 & 14.88 & 18.37 & 15.09 \\
OmniQuant*~\cite{shao2023omniquant} & 4/4 & 15.13 & 16.72 & 15.67 & 14.46 & 17.03 & 14.91 & 14.63 & 15.97 & 14.70 \\
RepQuant (ours) & 4/4 & \textbf{13.97} & \textbf{15.49} & \textbf{14.12} & \textbf{13.73}& \textbf{15.68} & \textbf{14.01} & \textbf{13.19} & \textbf{14.26} & \textbf{13.40} \\ 
\midrule
SmoothQuant*~\cite{xiao2023smoothquant} & 6/6 & 13.22 & 16.12 & 14.09 & 12.38 & 14.51 & 13.86 & 11.87 & 13.73 & 13.15 \\
Outlier Suppression+*~\cite{wei2023outlier} & 6/6 & 11.45 & 13.73 & 12.30 & 11.53 & 13.37 & 12.92 & 10.68 & 13.15 & 11.17 \\
RPTQ*~\cite{yuan2023rptq} & 6/6 & \textbf{11.20} & 13.95 & 12.08 & 11.02 & 14.41 & 11.78 & \textbf{10.30} & 14.80 & 11.82 \\
OmniQuant*~\cite{shao2023omniquant} & 6/6 & 11.34 & 13.87 & 12.19 & 10.92 & 13.24 & \textbf{11.61} & 10.61 & 12.91 & 11.33 \\
RepQuant (ours) & 6/6 & 11.23 & \textbf{13.61} & \textbf{12.04} & \textbf{10.82} & \textbf{12.95} & 11.70 & 10.47 & \textbf{12.57} & \textbf{11.10} \\ 
\bottomrule
\end{tabular}
 \begin{tablenotes}
        \footnotesize
        \item[$\dagger$] Note that ``*" denotes the reproduced results using open-source codes, with the aim of aligning quantization configurations to ensure fair comparisons. Please see Section \ref{sec:setup}-\emph{Comparison methods} for details of the settings.
      \end{tablenotes}
    \end{threeparttable}
\label{tab:opt}
\end{table*}
\begin{table*}[h]
\centering
\small
\setlength{\tabcolsep}{7pt}
\caption{Quantization results of LLaMA models on zero-shot tasks, including  PIQA,
ARC, HellaSwag, and Winogrande. We report the zero-shot accuracy on each task as well as the average accuracy. Here, ``Prec. (W/A)'' denotes the quantization precision.} 
\begin{tabular}{@{}ccccccccc@{}}
\toprule
\textbf{Model} & \textbf{Method} & \textbf{Prec. (W/A)} & \textbf{PIQA} & \textbf{ARC-e} & \textbf{Arc-c} & \textbf{HellaSwag} & \textbf{Winogrande} & \textbf{Average} \\ \midrule
\multirow{8}{*}{LLaMA-7B} & Full-Precision & 32/32 & 77.47 & 52.48 & 41.46 & 73.00 & 67.07 & 62.30 \\ \cmidrule(l){2-9} 
 & SmoothQuant*~\cite{xiao2023smoothquant} & 4/4 & 44.35 & 24.58 & 17.26 & 21.06 & 42.81 & 30.01 \\
 & OmniQuant*~\cite{shao2023omniquant} & 4/4 & 63.42 & 42.58 & 30.12 & 53.83 & 50.37 & 48.06 \\
 & RepQuant (ours) & 4/4 & \textbf{65.87} & \textbf{44.39} & \textbf{30.90} & \textbf{55.76} & \textbf{53.15} & \textbf{50.01} \\ \cmidrule(l){2-9} 
 & SmoothQuant*~\cite{xiao2023smoothquant} & 6/6 & 75.04 & 50.26 & 38.70 & 71.11 & 64.18 & 59.86 \\
 & OmniQuant*~\cite{shao2023omniquant} & 6/6 & 76.24 & 50.80 & 39.57 & 71.16 & 64.22 & 60.40 \\
 & RepQuant (ours) & 6/6 & \textbf{76.57} & \textbf{51.29} & \textbf{40.14} & \textbf{71.33} & \textbf{64.82} & \textbf{60.83} \\ \midrule
\multirow{8}{*}{LLaMA-13B} & Full-Precision & 32/32 & 79.10 & 59.89 & 44.45 & 76.21 & 70.31 & 65.99 \\ \cmidrule(l){2-9} 
 & SmoothQuant*~\cite{xiao2023smoothquant} & 4/4 & 57.23 & 34.87 & 28.61 & 48.47 & 49.78 & 43.79 \\
 & OmniQuant*~\cite{shao2023omniquant} & 4/4 & 66.17 & 45.07 & 31.46 & 55.73 & 53.19 & 50.32 \\
 & RepQuant (ours) & 4/4 & \textbf{68.46} & \textbf{46.71} & \textbf{32.38} & \textbf{57.71} & \textbf{54.38} & \textbf{51.93} \\ \cmidrule(l){2-9} 
 & SmoothQuant*~\cite{xiao2023smoothquant} & 6/6 & 76.11 & 54.46 & 39.85 & 71.87 & 66.27 & 61.71 \\
 & OmniQuant*~\cite{shao2023omniquant} & 6/6 & 76.52 & 55.80 & 40.08 & \textbf{73.32} & 66.52 & 62.45 \\
 & RepQuant (ours) & 6/6 & \textbf{77.37} & \textbf{57.26} & \textbf{41.60} & 73.25 & \textbf{67.33} & \textbf{63.36} \\\midrule
\multirow{8}{*}{LLaMA-65B} & Full-Precision & 32/32 & 80.79 & 58.71 & 46.24 & 80.72 & 77.50 & 68.79 \\ \cmidrule(l){2-9} 
 & SmoothQuant*~\cite{xiao2023smoothquant} & 4/4 & 61.77 & 36.94 & 27.62 & 35.69 & 50.74 & 42.55 \\
 & OmniQuant*~\cite{shao2023omniquant} & 4/4 & 69.44 & 45.10 & 31.84 & 61.77 & 55.62 & 52.75 \\
 & RepQuant (ours) & 4/4 & \textbf{70.76} & \textbf{46.86} & \textbf{34.07} & \textbf{64.52} & \textbf{57.93} & \textbf{54.83} \\ \cmidrule(l){2-9} 
 & SmoothQuant*~\cite{xiao2023smoothquant} & 6/6 & 79.31 & 56.74 & 44.59 & 79.03 & 74.20 & 66.77 \\
 & OmniQuant*~\cite{shao2023omniquant} & 6/6 & \textbf{79.87} & 57.18 & 44.93 & 78.90 & 74.79 & 67.13 \\
 & RepQuant (ours) & 6/6 & 79.69 & \textbf{57.96} & \textbf{45.31} & \textbf{78.99} & \textbf{75.48} & \textbf{67.49} \\ \bottomrule
\end{tabular}
\label{tab:llama}
\end{table*}

\subsection{Performance Evaluation on Language Transformers}
RepQuant is also comprehensively evaluated on large language models. In particular, we assess the perplexity of the OPT models on language generation tasks and examine the accuracy of the LLaMA models on zero-shot tasks. 

\textbf{Quantization results for OPT models}  
We quantize the noteworthy OPT model family, encompassing various scales such as 6.7B, 13B, and 30B. The perplexity results for language generation are reported in Table \ref{tab:opt}. Note that the quantization configurations of the existing methods are not aligned, and to ensure fair comparisons, we reproduce them at aligned settings based on open-source codes, which are marked by ``*" in the table.
The proposed RepQuant consistently demonstrates exceptional potential and marked performance superiority under identical settings. In the case of W4/A4 quantization, SmoothQuant, which manually smoothes outliers of LayerNorm activations before quantization, becomes impractical and results in prohibitively high perplexity levels. Although Outlier Suppression+ incorporates additional shifting, it still leads to a less optimal transformation, thus offering limited performance enhancement. The reorder operation in RPTQ shows some utility, albeit with the downside of increased inference overhead, and its performance is still far from satisfactory. Additionally, Omniquant overlooks static quantization of activations, resulting in a noticeable performance discrepancy compared to the full-precision baseline. Fortunately, our RepQuant effectively addresses these challenges, presenting a robust solution.
For instance, when quantizing OPT-13B using RepQuant, the perplexity on WIKI, PTB, and C4 datasets are 13.73, 15.68, and 14.01, respectively. Specifically for WIKI dataset, RepQuant achieves notable improvements, reducing the perplexity by 2.20, 2.01, and 0.73 when compared to the Outlier Suppression+, RPTQ, and Omniquant, respectively.

\begin{figure*}
	\centering
	\includegraphics[width=0.75\linewidth]{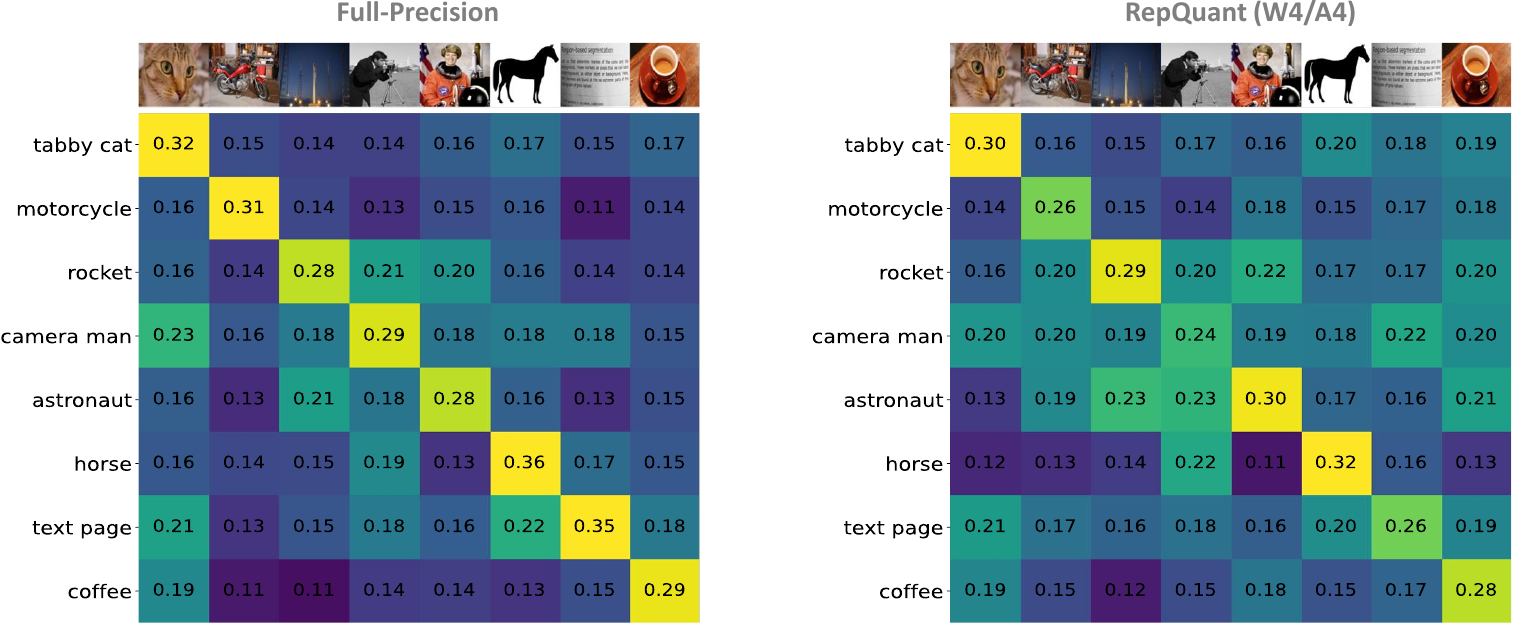}
	\caption{Visualization of cosine similarity of encoded features in full-precision and quantized CLIP models. Here, ViT-B and Transformer-63M serve as the image encoder and text encoder, respectively. It is intuitively demonstrated that when performing W4/A4 quantization, RepQuant provides a minimal impact on the image-text matching ability of the CLIP model and thus potentially maintains its zero-shot accuracy and robustness.}
	\label{fig:clip}
\end{figure*}
\begin{table*}[t]
\centering
\small
\setlength{\tabcolsep}{8pt}
\caption{Quantization results of zero-shot performance of CLIP models on CIFAR100 and ImageNet datasets. We reproduce different quantization methods as baselines for image encoder and text encoder, respectively. Here, ``Prec. (W/A)'' denotes the quantization precision, i.e., the weights and activations are quantized to W and A bits, respectively.}
\begin{tabular}{@{}ccccccc@{}}
\toprule
\multicolumn{2}{c}{\textbf{CLIP Model}} & \multicolumn{2}{c}{\textbf{Method}} & \multirow{2}{*}{\textbf{Prec. (W/A)}} & \multirow{2}{*}{\textbf{CIFAR100}} & \multirow{2}{*}{\textbf{ImageNet}} \\
\textbf{Image Encoder}          & \textbf{Text Encoder}      & \textbf{Image Encoder} & \textbf{Text Encoder} &  &  &  \\ \midrule
\multirow{8}{*}{ViT-B} & \multirow{8}{*}{Transformer-63M} 
&  Full-Precision    &  Full-Precision  &   32/32      & 68.7 & 68.6 \\ \cmidrule(l){3-7} 
&&  PTQ4ViT~\cite{yuan2021ptq4vit} &  SmoothQuant*~\cite{xiao2023smoothquant}     &   4/4        & 5.4 &  1.9 \\
&&  RepQ-ViT~\cite{li2023repq}     &  SmoothQuant*~\cite{xiao2023smoothquant}     &   4/4        & 35.4 &  32.7 \\
&&  RepQuant (ours)  &  RepQuant (ours) &   4/4        & \textbf{60.5} & \textbf{61.1}  \\\cmidrule(l){3-7} 
&&  PTQ4ViT~\cite{yuan2021ptq4vit} &  SmoothQuant*~\cite{xiao2023smoothquant}     &   6/6        & 64.6 &  64.2 \\
&&  RepQ-ViT~\cite{li2023repq}     &  SmoothQuant*~\cite{xiao2023smoothquant}     &   6/6        & 66.1 & 65.7  \\
&&  RepQuant (ours)  &  RepQuant (ours) &   6/6        & \textbf{67.1} & \textbf{66.5}  \\\midrule
\multirow{8}{*}{ViT-L} & \multirow{8}{*}{Transformer-63M} 
&  Full-Precision    &  Full-Precision  &   32/32      & 77.9 & 75.3 \\ \cmidrule(l){3-7} 
&&  PTQ4ViT~\cite{yuan2021ptq4vit} &  SmoothQuant*~\cite{xiao2023smoothquant}     &   4/4        & 10.3 & 3.9  \\
&&  RepQ-ViT~\cite{li2023repq}     &  SmoothQuant*~\cite{xiao2023smoothquant}     &   4/4        & 45.7 &  42.0 \\
&&  RepQuant (ours)  &  RepQuant (ours) &   4/4        & \textbf{69.8} &  \textbf{68.1} \\\cmidrule(l){3-7} 
&&  PTQ4ViT~\cite{yuan2021ptq4vit} &  SmoothQuant*~\cite{xiao2023smoothquant}     &   6/6        & 74.1 &  71.6 \\
&&  RepQ-ViT~\cite{li2023repq}     &  SmoothQuant*~\cite{xiao2023smoothquant}     &   6/6        & 75.2 &  73.2 \\
&&  RepQuant (ours)  &  RepQuant (ours) &   6/6        & \textbf{76.0} & \textbf{74.1}  \\ \bottomrule
\end{tabular}
\label{tab:clip}
\end{table*}

\textbf{Quantization results for LLaMA models}
Our evaluation of RepQuant extends to the LLaMA model family, encompassing various scales including 7B, 13B, and 65B. We meticulously present the accuracy results of these models on multiple zero-shot tasks in Table \ref{tab:llama}, which further demonstrates the effectiveness and advantages of RepQuant. In the case of the LLaMA-7B model, using RepQuant at W4/A4 precision, we observe a substantial improvement in average accuracy. It outperforms SmoothQuant by 20.00\% and OmniQuant by 1.95\%. In W6/A6 quantization, RepQuant achieves an impressive average accuracy of 60.83\%, which is a mere 1.47\% reduction from the full-precision baseline. As we scale up the model size, the robustness of RepQuant remains evident. With the LLaMA-65B model, RepQuant markedly surpasses existing quantization methods. In W4/A4 quantization, it achieves an improvement of 12.28\% over SmoothQuant and 2.08\% over OmniQuant. In addition, when performing W6/A6 quantization, RepQuant yields similar results to the full-precision baseline, showcasing its significant advantage in handling larger-scale models.

\subsection{Performance Evaluation on Multi-Modal Transformers}
In addition to pure vision transformers and language transformers, we also evaluate the quantization performance of RepQuant on multi-modal transformers.
A notable multi-modal transformer is the CLIP model, characterized by its dual-tower architecture comprising an image encoder and a text encoder. And existing methods designed for specific tasks face challenges in adaptability, particularly across different tasks. For instance, applying SmoothQuant to the image encoder of CLIP could potentially compromise accuracy, thus we need to reproduce separate methods for each of the two encoders as the comparison baseline. In contrast, our proposed RepQuant demonstrates remarkable compatibility and versatility for model variants across various scales on multiple tasks, which allows it to quantize both encoders.

Table \ref{tab:clip} reports the quantization results of zero-shot performance of CLIP models on CIFAR100 and ImageNet datasets.
We delve into a detailed comparison and analysis, taking ViT-B and Transformer-63M as encoders as an example. In W4/A4 quantization, employing PTQ4ViT and SmoothQuant for quantizing the image and text encoders respectively proves to be impractical, leading to significant accuracy loss. Substituting PTQ4ViT with RepQ-ViT yields a notable improvement, yet it still falls short of expectations. Under this configuration, the zero-shot accuracies on CIFAR100 and ImageNet show a decrease of 33.3\% and 35.9\% respectively, compared to the full-precision baseline. Fortunately, when both encoders are quantized using RepQuant, the zero-shot performance is further improved substantially, by encouraging 25.1\% and 28.4\%, respectively. Moreover, in W6/A6 quantization, RepQuant achieves comparable performance to the full-accuracy baseline, with only marginal differences of 1.6\% and 2.1\% in accuracies on CIFAR100 and ImageNet.

We also visualize the cosine similarity of encoded features in full-precision and quantized CLIP models using RepQuant, as shown in Fig. \ref{fig:clip}.
The intuitive demonstration in W4/A4 quantization underscores a vital aspect of RepQuant: it exerts minimal impact on the integral image-text matching ability of the CLIP model. This finding is particularly significant as it suggests that RepQuant can preserve the intrinsic strengths of the CLIP model, notably its zero-shot accuracy and robustness, which are crucial for effective multi-modal learning.

\subsection{Ablation Studies}

\begin{table}[t]
\centering
\small
\setlength{\tabcolsep}{6pt}
\caption{Ablation studies of different quantizers (W4/A4) for LayerNorm activations. Here, ``Efficiency'' represents the inference efficiency of the quantized model, and ``Accuracy'' denotes the Top-1 accuracy (\%) for DeiT on ImageNet and the average accuracy (\%) for LLaMA on zero-shot tasks.}
\begin{tabular}{cccc}
\toprule
\textbf{Model} & \textbf{Method} & \textbf{Efficiency} &\textbf{Accuracy} \\
\midrule
\multirow{4.5}{*}{DeiT-S} & Full-Precision & - & 79.85 \\
\cmidrule{2-4}
& Layer-Wise Quant. & $\checkmark$ & 39.55 \\
& Channel-Wise Quant. & $\times$ & \textbf{75.84} \\
& Scale Reparam (ours) & $\checkmark$ & 75.21 \\
\midrule
\multirow{4.5}{*}{LLaMA-13B} & Full-Precision & - & 65.99 \\
\cmidrule{2-4}
& Layer-Wise Quant. & $\checkmark$ & 32.61 \\
& Channel-Wise Quant. & $\times$ & \textbf{52.38} \\
& Scale Reparam (ours) & $\checkmark$ & 51.93 \\
\bottomrule

\end{tabular}
\label{tab:ablation_1}
\end{table}

\begin{table}[t]
\centering
\small
\setlength{\tabcolsep}{6pt}
\caption{Ablation studies of different quantizers (W4/A4) for Softmax activations. Here, ``Efficiency'' represents the inference efficiency of the quantized model, and ``Accuracy'' denotes the Top-1 accuracy (\%) for DeiT on ImageNet and the average accuracy (\%) for LLaMA on zero-shot tasks.}
\begin{tabular}{cccc}
\toprule
\textbf{Model} & \textbf{Method} & \textbf{Efficiency} &\textbf{Accuracy} \\
\midrule
\multirow{4.5}{*}{DeiT-S} & Full-Precision & - & 79.85 \\
\cmidrule{2-4}
& Log2 Quant. & $\checkmark$ & 73.95 \\
& Log$\sqrt{2}$ Quant. & $\times$ & \textbf{75.21} \\
& Scale Reparam (ours) & $\checkmark$ & \textbf{75.21} \\
\midrule
\multirow{4.5}{*}{LLaMA-13B} & Full-Precision & - & 65.99 \\
\cmidrule{2-4}
& Log2 Quant. & $\checkmark$ & 50.33 \\
& Log$\sqrt{2}$ Quant. & $\times$ & \textbf{51.93} \\
& Scale Reparam (ours) & $\checkmark$ & \textbf{51.93} \\
\bottomrule

\end{tabular}
\label{tab:ablation_2}
\end{table}

\begin{table}[t]
\centering
\small
\setlength{\tabcolsep}{10pt}
\caption{Ablation studies of the effect of different modules on quantization performance (W4/A4). Here, ``Accuracy'' denotes the Top-1 accuracy (\%) for DeiT on ImageNet and the average accuracy (\%) for LLaMA on zero-shot tasks.}
\begin{tabular}{cccc}
\toprule
\textbf{Model}                   & \textbf{Dual Clipping} & \textbf{GPTQ} & \textbf{Accuracy} \\ \midrule
\multirow{4}{*}{DeiT-S} &   $\times$     &  $\times$ &   69.03   \\
                        &    $\checkmark$     &  $\times$ &   71.56   \\
                        &    $\times$     &  $\checkmark$  &   72.60   \\
                        &   $\checkmark$   &  $\checkmark$ &  \textbf{75.21} \\\midrule
\multirow{4}{*}{LLaMA-13B} &   $\times$     &  $\times$ &    47.35  \\
                        &    $\checkmark$     &  $\times$ &  49.52 \\
                        &    $\times$     &  $\checkmark$  & 49.68 \\
                        &   $\checkmark$   &  $\checkmark$ &  \textbf{51.93}\\ \bottomrule
\end{tabular}
\label{tab:ablation_3}
\end{table}

\begin{table}[t]
\centering
\small
\setlength{\tabcolsep}{6pt}
\caption{Ablation studies of different contraction methods in per-channel dual clipping (W4/A4). Here, ``Accuracy'' denotes the Top-1 accuracy (\%) for DeiT on ImageNet and the average accuracy (\%) for LLaMA on zero-shot tasks, and ``GPU Sec." is the runtime for processing one layer on one A6000 GPU.}
\begin{tabular}{cccc}
\toprule
\textbf{Model} & \textbf{Method} & \textbf{GPU Sec.} &\textbf{Accuracy} \\
\midrule
\multirow{4.5}{*}{DeiT-S} & Full-Precision & - & 79.85 \\
\cmidrule{2-4}
& Search-Based~\cite{tu2023toward} & 495 & 74.69 \\
& Numerical Direct & 16 & 72.83 \\
& Sigmoid (ours) & 17 & \textbf{75.21} \\
\midrule
\multirow{4.5}{*}{LLaMA-13B} & Full-Precision & - & 65.99 \\
\cmidrule{2-4}
& Search-Based~\cite{tu2023toward} & 5100 & 50.87 \\
& Numerical Direct & 44 & 49.11 \\
& Sigmoid (ours) & 46 & \textbf{51.93} \\
\bottomrule

\end{tabular}
\label{tab:ablation_4}
\end{table}

\textbf{Quantizers for LayerNorm activations}
Table \ref{tab:ablation_1} presents the results of different quantizers (at W4/A4 precision) for post-LayerNorm activations. Using DeiT-S as a representative case, layer-wise quantization does not effectively capture the data distribution, leading to a modest accuracy of 39.55\%. In contrast, channel-wise quantization addresses this limitation, significantly improving accuracy to 75.84\%. However, this approach does not align well with hardware requirements for efficient inference. Fortunately, our scale reparameterization method bridges the gap by converting channel-wise quantization to layer-wise quantization, thereby facilitating both high accuracy and efficient inference. Since this process slightly alters the distribution in the subsequent linear layer, there is a minor impact on the accuracy of GPTQ, resulting in a marginal decrease (0.63\%) in overall performance compared to channel-wise quantization.

\textbf{Quantizers for Softmax activations}
Table \ref{tab:ablation_2} presents the results of different quantizers (at W4/A4 precision) for post-Softmax activations. The log$\sqrt{2}$ quantizer is notably effective in fitting the highly skewed distributions of attention scores, particularly excelling in representing dispersed larger values. This results in a significant accuracy improvement compared to the basic log2 quantizer, as evidenced in both DeiT-S (1.26\%) and LLaMA-13B (1.60\%) cases. To address the efficiency shortcomings of log$\sqrt{2}$ quantizers, our scale reparameterization method facilitates their conversion to log2 quantizers. Notably, for post-Softmax activations, the scale reparameterization is a mathematically equivalent transformation, thus maintaining the same level of accuracy as achieved with log$\sqrt{2}$ quantizers, with only a marginal increase in computational overhead during inference.

\textbf{Effects of per-channel dual clipping and GPTQ}
The individual impacts of the new modules incorporated in RepQuant, per-channel dual clipping and GPTQ, are examined distinctly, with the results detailed in Table \ref{tab:ablation_3}. In the W4/A4 quantization for DeiT-S, the absence of both modules, aligning with the original RepQ-ViT configuration, results in a substantial accuracy deficit of 10.82\% compared to the full precision baseline. The per-channel dual clipping technique, adept at detecting and eliminating outliers in LayerNorm activations at a fine-grained level, enhances accuracy by 2.53\%. Meanwhile, GPTQ, which is integrated into the sequential pipeline through the reconstruction of quantized weights, contributes a 3.57\% boost in performance. Notably, the benefits of these two modules are cumulative, collectively elevating the accuracy to 75.21\%. This pattern of improvement is similarly observable in the LLaMA-13B model, underscoring the effectiveness of the two modules.

\textbf{Contraction methods in per-channel dual clipping}
We evaluate different contraction methods in per-channel dual clipping, as shown in Table \ref{tab:ablation_4}. The search-based method becomes impractical for the fine-grained case, because it takes considerable time to perform a single search for each channel, especially for large language models (e.g., LLaMA-13B). For the learning-based method, the numerical direct method, which takes the upper and lower bound values as the optimization objective, leads to inaccurate and unstable optimization, and even leads to negative optimization (0.57\%) for the LLaMA-13B model. On the contrary, our Sigmoid-based method demonstrates a more accurate and stable approach for dual clipping, which yields a performance improvement of 1.06\% and 2.82\% over the search-based and numerical direct methods, respectively, for the LLaMA-13B model.

\section{Conclusions}
In this paper, we propose RepQuant, an innovative post-training quantization framework for large transformer models, built on top of RepQ-ViT. Our approach adopts the quantization-inference decoupling paradigm, utilizing complex quantizers in the inference process, and simple and hardware-compatible quantizers in the inference process. The two processes are explicitly bridged via scale reparameterization, which is a mathematically equivalent transformation. Specifically, RepQaunt tackles the extreme distributions of two components: for post-LayerNorm activations with severe inter-channel variation, we initially apply channel-wise quantization and then reparameterize it to layer-wise quantization; for post-Softmax activations with power-law features, we initially apply log$\sqrt{2}$ quantization and then reparameterize it to log2 quantization. Furthermore, we incorporate a learnable per-channel dual clipping technique, adept at efficiently identifying outliers in the unbalanced activations at a fine-grained level. We also seamlessly integrate the concept of quantized weight reconstruction into the above procedure, aiming to further push the performance limits.
We conduct comprehensive experiments across various large-scale transformer variants, encompassing a range of tasks in vision, language, and multi-modal fields. Encouragingly, our RepQuant framework showcases substantial performance benefits and robust capabilities in these diverse applications.

\bibliographystyle{ieeetr}
\bibliography{egbib}




\end{document}